\documentclass[10pt]{article} % For LaTeX2e

\usepackage[preprint]{rlj} % Should be uncommented for preprint versions

\usepackage{amssymb}            % Defines common symbols like \mathbb R
\usepackage{mathtools}          % Extends amsmath, providing common math tools
\usepackage{mathrsfs}           % Enables \mathscr, which can work in cases that \mathcal does not
\usepackage{graphicx}           % For including images
\usepackage[space]{grffile}     % For spaces in image names
\usepackage{url}                % For displaying URLs
\usepackage{lipsum}             % For placeholder text
\usepackage{algorithm}
\usepackage{algorithmic}
\usepackage{tikz}
\usepackage{microtype}
\usepackage{subfigure}
\usepackage{booktabs} % for professional tables
\usepackage{hyperref}

\usepackage{amsmath}
\usepackage{bm}
\usepackage{amssymb}
\usepackage{mathtools}
\usepackage{amsthm}

\usepackage[capitalize,noabbrev]{cleveref}

\theoremstyle{plain}
\newtheorem{theorem}{Theorem}[section]

\newtheorem{lemma}[theorem]{Lemma}

\theoremstyle{definition}
\newtheorem{definition}[theorem]{Definition}
\newtheorem{assumption}[theorem]{Assumption}
\theoremstyle{remark}

\title{Non-Stationary Latent Auto-Regressive Bandits}

\setrunningtitle{Non-Stationary Latent Auto-Regressive Bandits}

\author{Anna L. Trella\textsuperscript{1}, Walter Dempsey\textsuperscript{2}, Asim H. Gazi\textsuperscript{1}, Ziping Xu\textsuperscript{1}, Finale Doshi-Velez\textsuperscript{1}, Susan A. Murphy\textsuperscript{1}
}

\emails{annatrella@g.harvard.edu, wdem@umich.edu, agazi@seas.harvard.edu, zipingxu@fas.harvard.edu, finale@seas.harvard.edu, samurphy@fas.harvard.edu}

\affiliations{
$^{1}$\textbf{School of Engineering and Applied Sciences, Harvard University, Cambridge, MA USA}\\
$^{2}$\textbf{Department of Biostatistics, University of Michigan, Ann Arbor, MI USA}
}

\contribution{
    This paper introduces Latent AR LinUCB (LARL), an efficient online algorithm designed for non-stationary MABs where the non-stationarity is due to a latent, auto-regressive (AR) state. LARL forms good predictions of reward means by implicitly predicting the latent state using past rewards and actions. This strategy can be seen as an approximation of a steady-state Kalman filter with ground-truth system parameters. 
    }
    {
    The setting we consider is motivated by real-world applications where the non-stationary mechanism can be modeled by a latent state, but there is no budget for the non-stationarity.
    Existing approaches that consider similar settings rely on the latent state being discrete \citep{hong2020non, nelson2022linearizing} or require knowing the ground truth parameters or quality historical data to recover parameters \citep{liu2023nonstationary, chen2024non}.
    }

\contribution{
    We present an interpretable regret bound for LARL against the dynamic oracle. The regret bound allows practitioners to interpret the performance of LARL with respect to the level of non-stationarity in the environment and the complexity of learning parameters online.
    }
    {
    Sub-linear regret with respect to the dynamic oracle is only possible in environments with a budget for non-stationarity that is sub-linear in $T$. For example, \citet{besbes2014stochastic} assume a finite constant (variation budget) of how much the mean rewards can change over time and \citet{garivier2011upper} assume a finite number of changes to the mean reward.
    We show that in our setting, LARL achieves sub-linear regret if the noise variance on the latent state process is sufficiently small with respect to $T$.
    }

\contribution{
    We demonstrate that LARL can outperform (achieve lower regret) against various stationary and non-stationary baselines in the non-stationary bandit environment where reward means change due to a latent AR state.
    }
    {
    We consider cumulative regret across time and pairwise comparisons of methods in terms of total cumulative regret. To offer a fair comparison, baseline methods were only considered if they implemented an online learning strategy.
    }

 % Your keywords

\summary{
For the non-stationary multi-armed bandit (MAB) problem, many existing methods allow a general mechanism for the non-stationarity, but rely on a budget for the non-stationarity that is sub-linear to the total number of time steps $T$.
In many real-world settings, however, the mechanism for the non-stationarity can be modeled, but there is no budget for the non-stationarity.
We instead consider the non-stationary bandit problem where the reward means change due to a latent, auto-regressive (AR) state.
We develop Latent AR LinUCB (LARL), an online linear contextual bandit algorithm that does not rely on the non-stationary budget, but instead forms good predictions of reward means by implicitly predicting the latent state.
The key idea is to reduce the problem to a linear dynamical system which can be solved as a linear contextual bandit.
In fact, LARL approximates a steady-state Kalman filter and efficiently learns system parameters online. 
We provide an interpretable regret bound for LARL with respect to the level of non-stationarity in the environment.
LARL achieves sub-linear regret in this setting if the noise variance of the latent state process is sufficiently small with respect to $T$.
Empirically, LARL outperforms various baseline methods in this non-stationary bandit problem.

}

\begin{document}

\makeCover  % Create the cover page
\maketitle  % Make the title section

\begin{abstract}
    For the non-stationary multi-armed bandit (MAB) problem, many existing methods allow a general mechanism for the non-stationarity, but rely on a budget for the non-stationarity that is sub-linear to the total number of time steps $T$.
% This assumption is violated in many real-world settings, especially in emerging scientific fields such as behavioral health or education where there are few mechanistic models of the environment. 
In many real-world settings, however, the mechanism for the non-stationarity can be modeled, but there is no budget for the non-stationarity.
We instead consider the non-stationary bandit problem where the reward means change due to a latent, auto-regressive (AR) state.
% We prove that this non-stationary MAB can be rewritten as a linear contextual bandit with features formed using past history of rewards and actions. 
We develop Latent AR LinUCB (LARL), an online linear contextual bandit algorithm that does not rely on the non-stationary budget, but instead forms good predictions of reward means by implicitly predicting the latent state.
% exploiting the structure of the latent AR state.
% uses past history to predict the latent state and to implicitly learn parameters online. 
The key idea is to reduce the problem to a linear dynamical system which can be solved as a linear contextual bandit.
In fact, LARL approximates a steady-state Kalman filter and efficiently learns system parameters online. 
% Our algorithm requires no knowledge of $k$.
We provide an interpretable regret bound for LARL with respect to the level of non-stationarity in the environment.
LARL achieves sub-linear regret in this setting if the noise variance of the latent state process is sufficiently small with respect to $T$.
Empirically, LARL outperforms various baseline methods in this non-stationary bandit problem.

\end{abstract}

\section{Introduction}
\label{sec_intro}

% what is a non-stationary bandit
% who are people who have studied it
% what is our new non-stationary formulation?
% why is it important to consider our new formulation?
% what are some examples of our new formulation?

%%% what is the take home message? %%%
% We aim to provide guidance for real-world domains such as behavioral health and education  where practitioners notice signs of non-stationarity. In order to accomplish this, wedevelop an approach that is able to handle realistic sources of non-stationarity. 
In the classical formulation of the stochastic multi-armed bandit (MAB) problem \citep{lattimore2020bandit}, the rewards are assumed to be independently and identically drawn from a fixed distribution. In the non-stationary formulation \citep{auer2002nonstochastic}, the reward means, instead, change over time. 
While many existing approaches for non-stationary bandits allow an arbitrary mechanism for the non-stationarity, they rely on some budget to the non-stationarity that is sub-linear to the total number of time steps $T$. For example, in \cite{besbes2014stochastic} there is a variation budget for the amount of change in the mean rewards, and in \cite{garivier2011upper}
% there are epochs for which the reward means stay fixed and 
there is a budget for the number of changes.
% Why are we interested in this setup? This non-stationary environment is what we expect in the clinical setting we care about. We think the latent state is temporally correlated.
In contrast, for many real-world applications, the non-stationarity mechanism can be modeled as a latent state with temporal dependencies, but with restless non-stationarity. 
% REAL WORLD EXAMPLE %
% \alt{Here is the example I came up with but would love feedback on a better mobile health example!}
For example, in mobile health applications, bandit algorithms are used to optimize notifications to maximize users' health outcomes (rewards). User burden from using the app is an evolving latent process with temporal dependencies and can cause the health outcome to decline over time (non-stationarity). 
% Previous studies in digital health found that users can be burdened when receiving too many notifications \citep{mcelfresh2020matching}. Therefore scientists in these fields can incorporate recent notification history into the context to infer user burden \citep{trella2023reward}.

Motivated by these real-world settings, we study a non-stationary bandit problem with a realistic source of non-stationarity. In this problem, changes in the mean reward of the arms over time are due to some latent, auto-regressive (AR) state of order $k$. This problem is represented by the graphical model in Figure~\ref{fig_latent_ar_dag}.
Such a latent state causes smooth changes to the mean rewards as opposed to abrupt changes; however, the variation budget or the budget on the number of changes could scale linearly with $T$.

Our approach to solving the non-stationary bandit problem in Figure~\ref{fig_latent_ar_dag} leverages the graphical structure and reduces the problem instead to the well-studied problem of linear dynamical systems (Section~\ref{sec_connect_w_lds}); we then show that the linear dynamical system can be solved as a linear contextual bandit (Section~\ref{sec_reduction}). By leveraging the structure of the non-stationarity, we can offer a finer theoretical analysis and design a more specific algorithm that can outperform general non-stationary algorithms.
% \alt{Chat with Ziping notes.
% In our setting we are assuming a specific structure for the non-stationarity and this helps us do a finer analysis and inspires a specific algorithm that outperforms general non-stationary algorithms because we leverage the structure of the problem. We validate this improvement in simulation studies.
% }
% with a temporal autocorrelation structure.
% We want an implementable algorithm that can be run in real-life.

\textbf{Contributions.}
 We propose Latent AR LinUCB or LARL (Algorithm~\ref{alg_latent_ar_ucb}), an 
 % computationally-efficient 
 online linear contextual bandit algorithm that maintains good reward predictions by using past history to predict the current latent state and to learn parameters online.
 The reward model maintained by LARL can be seen as an approximation to the steady-state Kalman filter with access to ground-truth system parameters.
We present an interpretable regret bound for LARL against the dynamic oracle (Theorem~\ref{thm_regret}). In our setting, LARL achieves sub-linear regret if the noise variance of the latent state process is sufficiently small with respect to the total number of time steps $T$.
 We validate in simulation studies (Section~\ref{sec_experiemnts}) that LARL outperforms various baseline methods in the non-stationary latent AR environment.

\section{Related Works}
%% AR Book: http://repository.cinec.edu/bitstream/cinec20/1228/1/2008_Book_TimeSeriesAnalysis.pdf

% \subsection{Linear Dynamical Systems}
% \alt{should we add a RW section on filtering and LDS?}

\subsection{Non-Stationary Bandits}
Non-stationary bandits \citep{auer2002nonstochastic} extend the standard bandit problem to one with reward means changing over time. 
% While keeping track of previous observations may decrease the variance of mean reward estimates in standard settings, this historical information could be less relevant in non-stationary settings due to the possible changes in the reward means.
% In a standard, stationary setting, keeping track of previous observations may decrease the variance of mean reward estimates. However, the non-stationary environment implies that this historical information could be less relevant due to the possible changes in the underlying rewards. 
% https://arxiv.org/pdf/2305.10718.pdf is a good resource for explaining different non-stat MABs
Many approaches have been proposed for non-stationary bandits including change point detection \citep{mellor2013thompson}, sliding window \citep{garivier2011upper, cheung2019learning, trovo2020sliding}, restarting \citep{besbes2014stochastic, viappiani2013thompson}, and discounting the effect of past observations \citep{kocsis2006discounted, garivier2011upper, raj2017taming}.
A majority of these methods were developed for an arbitrary mechanism for controlling the non-stationarity, but rely on a budget for the amount of changes that is sub-linear in $T$. For example, there is a total budget for the amount of change \citep{besbes2014stochastic} or to the number of changes \citep{garivier2011upper}.
In contrast, our setting has a specific mechanism controlling the non-stationarity (i.e., a latent AR state), but the budget for the non-stationarity could scale linearly with $T$.
Others have also formulated non-stationarity through a latent state. These works propose methods for maintaining a posterior belief over the latent state and acting according to it \citep{hong2020non, nelson2022linearizing}.
\citet{nelson2022linearizing} rely on the latent state being discrete, as the dimension of the linear contextual bandit is the cardinality of the set of latent state values. 
% each entry in the context is the posterior probability $p(z_t = z | \mathcal{H}_{t - 1})$. 
This approach is not applicable in our setting where the latent state is continuous.
\citet{hong2020non} use particle filtering to sample from the joint posterior of the latent state and model parameters. 
% Each particle represents a possible state or parameter value. When the dimensionality of the state or parameter space increases, the number of particles needed to adequately cover the space also increases. In higher dimensions, the number of particles needed to maintain an accurate approximation grows exponentially, which leads to higher computational costs.
However, their approach would be computationally challenging in our setting, as the number of particles needed increases with the number of latent state values. 
% However, because they assume the latent state is discrete, the posterior distribution can be easily parameterized by the number of latent states, which is not applicable to our setting where the latent state is continuous.
Also similar to our work is \citet{gornet2022stochastic} which implements a similar reduction to a linear contextual bandit; however, their algorithm heavily relies on exogenous context to predict the latent state, which is not present in our setting.

\subsection{AR Bandits}
Autoregressive (AR) processes \citep{brockwell2009time} are studied extensively across many fields to model temporal dependencies in many real-world processes.
Some bandit formulations involve rewards evolving auto-regressively, rather than a latent state, with fixed reward means and known AR order $k$ \citep{bacchiocchi2022autoregressive}. 
Due to the fixed reward means, these settings are \textit{stationary} bandits, which allows for the direct application of standard linear bandit theory.
% This is not possible in our non-stationary setting, where the reward mean changes. Past reward realizations instead help us infer the latent state, which in turn gives us information on the current mean reward for each action.
While some have proposed similar settings where the non-stationarity is dictated by an AR process \citep{liu2023nonstationary, chen2024non}, these papers assume either access to ground-truth parameters or quality offline data to learn these parameters. Instead, we are interested in learning AR and reward parameters online. Furthermore, the predictive sampling method in \citet{liu2023nonstationary} is developed for the Bayesian framework and their main goal is to compare to traditional Thompson sampling. 
At first glance, one may notice that the AR(1) setting presented in \citet{chen2024non} is very similar. However, their method relies on a stronger assumption that the agent observes the true mean reward for the same action taken at the previous time step. In our setting, the agent never observes the exact mean reward for any action.

\section{Problem Setting}
\subsection{Notation}
We introduce the following notation used throughout the paper.
For some vector $v \in \mathbb{R}^d$, we use $\|v\| = \sqrt{\langle v, v\rangle}$ to denote the L2-norm of $v$ and $v^{\top}$ to denote the transpose of $v$. $\mathbf{1}$ denotes a vector of all 1s and $e_j$ denotes the standard basis vector with 1 in the $j$th component and 0s elsewhere.
For a square matrix $M \in \mathbb{R}^{d \times d}$, $M^{-1}$ denotes the inverse of $M$. 
$\lambda_{\max}(M)$ is the largest eigenvalue of $M$.
$\|M\|_{\text{op}} = \sigma_{\max}(M)$ is the operator norm or largest singular value of $M$.  
If $M$ is positive semi-definite (PSD), then 
$M^{1/2}$ denotes the square root of $M$ such that $M^{1/2}M^{1/2} = M$, 
and $\|v\|_M^2 = v^{\top} M v$ denotes the square of the weighted L2-norm of $v$. 
We use $[i, j]$ to denote the set of positive integers from $i$ to $j$, inclusive.
% $\Tilde{O}(\cdot)$ denotes big $O$ notation that ignores logarithm factors.
% Walter says when we introduce big O notation, we are considering a sequence of $T$ where $T$ grows large.
% $R_{1:\tau}$ denotes the set of reward realizations from \textit{all} actions up to and including time step $\tau$. $r_{1:\tau} = \{r_1, r_2,...,r_{\tau}\}$ denotes the set of realized rewards corresponding to the single selected action at each time step up to and including $\tau$.
$\mathbb{I}[\cdot]$ denotes the indicator function.  %For random variables $x, y$, we use $x \overset{d}{=} y$ to denote that $x$ and $y$ have the same distribution.
$\mathcal{H}_{t - 1}$ is the entire history of information observed up to, but not including, time $t$.
\begin{figure*}
  \centering
  \begin{tikzpicture}[
    node distance=1.4cm, % Adjust the distance between nodes
    every node/.style={draw, circle, minimum size=0.8cm, inner sep=0pt},
    shaded/.style={fill=gray!30}, % Style for shaded nodes
    invisible/.style={circle, draw=white, minimum size=0cm},
  ]
    % Define nodes with circles around them
    \node (A) {$z_1$};
    \node (B) [right of=A, invisible] {$\cdots$};
    \node (C) [below of=A, shaded] {$r_1$};
    \node (D) [below of=C, shaded] {$a_1$};

    \node (E) [right of=B]{$z_{t - k}$};
    \node (F) [right of=E, invisible] {$\cdots$};
    \node (G) [below of=E, shaded] {$r_{t - k}$};
    \node (H) [below of=G, shaded] {$a_{t - k}$};

    \node (I) [right of=F]{$z_{t - 1}$};
    \node (J) [right of=I, invisible] {$\cdots$};
    \node (K) [below of=I, shaded] {$r_{t - 1}$};
    \node (L) [below of=K, shaded] {$a_{t - 1}$};

    \node (M) [right of=J]{$z_t$};
    \node (N) [right of=M, invisible] {$\cdots$};
    \node (O) [below of=M, shaded] {$r_t$};
    \node (P) [below of=O, shaded] {$a_t$};
    %% last column
    \node (Q) [right of=N]{$z_T$};
    \node (R) [below of=Q, shaded] {$r_T$};
    \node (S) [below of=R, shaded] {$a_T$};

    % Draw directed edges
    \draw[->, line width=1.0pt] (A) -- (B); % Increase the arrow length at the start
    \draw[->, line width=1.0pt] (A) -- (C);
    \draw[->, line width=1.0pt] (D) -- (C);
    \draw[->, line width=1.0pt] (B) -- (E);

    % z_{t - k}
    \draw[->, line width=1.0pt] (E) -- (F); % Increase the arrow length at the start
    \draw[->, line width=1.0pt] (E) -- (G);
    \draw[->, line width=1.0pt] (H) -- (G);
    \draw[->, line width=1.0pt] (F) -- (I);
    % from z_{t - k} to z_{t - 1}
    \draw[->, line width=1.0pt] (E) to[out=90, in=90] (I);
    % from z_{t - k} to z_{t}
    \draw[->, line width=1.0pt] (E) to[out=90, in=90] (M);
    
    % z_{t - 1}
    \draw[->, line width=1.0pt] (I) -- (J); 
    \draw[->, line width=1.0pt] (I) -- (K);
    \draw[->, line width=1.0pt] (L) -- (K);

    % z_t
    \draw[->, line width=1.0pt] (J) -- (M);
    \draw[->, line width=1.0pt] (M) -- (O);
    \draw[->, line width=1.0pt] (P) -- (O);
    \draw[->, line width=1.0pt] (M) -- (N);
    
    \draw[->, line width=1.0pt] (N) -- (Q);
    \draw[->, line width=1.0pt] (Q) -- (R);
    \draw[->, line width=1.0pt] (S) -- (R);
    
  \end{tikzpicture}
  \caption{Graphical Model for Non-Stationary Latent Auto-regressive Bandits.
  % this graph shows that past rewards can be used to infer z_t which gives me information about r_t. But past rewards do not directly affect / cause r_t.
  }
  \label{fig_latent_ar_dag}
\end{figure*}

\subsection{Non-Stationary Latent Auto-regressive Bandits}
\label{latent_ar_setting}
%%%%% NON-STATIONARITY SETTING %%%%%%
We consider a non-stationary multi-armed bandit environment (Definition~\ref{def_non_stat_latent_auto_bandit}) where the true underlying reward depends on some latent state $z_t \in \mathbb{R}$ that evolves according to an auto-regressive (AR) process of order at most $k$. See Figure~\ref{fig_latent_ar_dag} for the graphical structure. 
% \citet{bacchiocchi2022autoregressive} presents a setting where the reward evolves auto-regressively; however, since the authors assume a fixed reward mean and known AR order $k$, this setting is stationary. Notice that in contrast, the reward being influenced by a latent AR process in our setting makes the reward distribution non-stationary. \alt{TODO: a lot of ICML reviewers got confused and thought that Bacchiocchi has already proposed our setting and solved it. Which is NOT TRUE! They said: ``theoretical analysis is quite straightforward, given the works on linear bandits of Abbasi-Yadkori et al. 2011 and the results on a similar setting of Bacchiocchi et al., 2022." I wrote an attempt below but I am having trouble making it more obvious that we are MUCH harder than bacchiocchi.} For the setting in \citet{bacchiocchi2022autoregressive}, one can use the past k reward realizations to form the current context which makes their setting a standard linear bandit. In our non-stationary setting, the past k reward realizations help us infer the latent state, which in turn gives us information for the current mean reward for each action.

\begin{definition}
    \label{def_non_stat_latent_auto_bandit}
    (Non-Stationary Latent Auto-regressive Bandit) 
    Let $\mathcal{A} \subset \mathbb{N}$ be the action space and initial latent states $[z_0,...,z_{k - 1}] \sim \mathcal{N}_k(\mu_0, \Sigma_0)$.
    % Let $\mathcal{A}$ be an arbitrary action space. 
    %Given the mean of the initial state $z_0$ to be $\mathbb{E}[z_0]= \mu_0$. 
    The interaction between the environment and the agent is as follows. For every time step $t \in [k, T]$:
    \begin{enumerate}
        \item The environment generates latent state $z_t$ of the form:
        \begin{equation}
        \label{latent_state}
            z_t = \gamma_0 + \sum_{j = 1}^k \gamma_jz_{t - j} + \xi_t, \;\;\; \xi_t \overset{\text{i.i.d.}}{\sim} \mathcal{N}(0, \sigma_z^2),
        \end{equation}
        where $\gamma_0, \gamma_1, ..., \gamma_k \in \mathbb{R}$. %The mean of the initial state $z_0$ is $\mathbb{E}[z_0]= \mu_0$.
        \item The agent selects action $a_t \in \mathcal{A}$ without observing $z_t$.
        \item %%% LINEAR REWARD %%%
        The environment then generates reward $r_t$ given latent state $z_t$ and action $a_t$, $r_t(a_t)$, where:
        \begin{align}
        \label{linear_reward}
            r_t(a) = \mu_a + \beta_a z_t + \epsilon_t(a), \;\; \epsilon_t(a) \overset{\text{i.i.d.}}{\sim} \mathcal{N}(0, \beta_a^2\sigma_r^2)
        \end{align}
        where $\mu_a, \beta_a, \epsilon_t(a) \in \mathbb{R}$ depends on the action $a$ and $\epsilon_t(a)$ is independent across actions and time steps.
        \item The agent observes $r_t$.
    \end{enumerate}
\end{definition}
Notice that in our setting, $z_t$ is not impacted by the action selected in the previous time step.
Also, notice that in Equation~\ref{linear_reward} the noise variance of the reward has an exact structure. This structure is needed to simplify the reduction in Lemma~\ref{linear_dynamic_system}.
% for the exact reduction present in Section BLAH.
But in practice, the algorithm (Algorithm~\ref{alg_latent_ar_ucb}) we present later does not require this noise structure.
% this is a special case of POMPDPs where the action taken does not impact the dynamics of the environment
To solve this non-stationary bandit problem, a natural approach is to form good predictions of $z_t$ and therefore good predictions of the reward means for each action. However, with no exogenous context available, the only observations one is given is the current history $\mathcal{H}_{t - 1}$ consisting of past actions and rewards. We will see in later sections that the method we develop can still perform well in such an environment, despite having limited information.

% Notice that because the latent state noise $\xi_t$ is non-vanishing, the non-stationarity in our problem is also non-vanishing. Equivalently, we can interpret our non-stationary formulation as one where the budget on the non-stationarity scales linearly in $T$.

% In order to simplify the analysis, we state our results assuming that the agent knows the noise variances $\sigma_z^2, \sigma_r^2$. However, one can relax these assumptions as for implementation, one can use past data to fit $\sigma_z^2, \sigma_r^2$. Notice that our method in Section~\ref{sec_alg} does not require knowledge of the AR order $k$.
 
% In order to simplify the analysis, we state our results assuming that the agent knows the noise variances $\sigma_z^2, \sigma_r^2$ and the AR order $k$. However, one can relax these assumptions as for implementation, one can use past data to fit $\sigma_z^2, \sigma_r^2$ and in Section \alt{BLAH}, we show $k$ can be learned via \alt{BLAH}.

\subsection{Connecting Latent AR Bandits With Linear Dynamical Systems}
\label{sec_connect_w_lds}
To assist with the reduction to a linear contextual bandit in Section~\ref{sec_reduction}, we first show that the latent AR bandit environment in Definition~\ref{def_non_stat_latent_auto_bandit} is a specific case of a linear dynamic system (LDS) with Gaussian noise. See Appendix~\ref{sec_lds_review} for a review of linear dynamical systems.
% We exploit the linear relationship in the AR process and in the reward function.

% \ziping{Anna, my suggestion is that you may skip some notations and simply say that check the appendix for the exact forms, as long as they are not critical for reporting the main theorem. }
% \alt{ANNA TODO: figure out which notation to keep here and which to put in the Appendix.}

\begin{lemma}
\label{linear_dynamic_system}
\textbf{(Linear Dynamical System)} % Linear Dynamic System
% Suppose latent state $z_t$ evolves in an $k$-order auto-regressive process of Equation~\ref{latent_state} with some initial state $\textbf{z}_0 \sim \mathcal{N}_k(\bm{\mu}_0, \Sigma_0)$ and for fixed action $a$, $r_t$ is defined in Equation~\ref{linear_reward}. 
The latent state process (Equation~\ref{latent_state}) and the reward function (Equation~\ref{linear_reward}) in Definition~\ref{def_non_stat_latent_auto_bandit} form a special case of a linear dynamical system with Gaussian noise. The system has state vector $\vec{z}_t \in \mathbb{R}^{k}$ which incorporates the most recent $k$ latent state realizations and measurement $y_{t} = \frac{r_{t} - \mu_{a_{t}}}{\beta_{a_{t}}} \in \mathbb{R}$.

\begin{align}
\label{state_evolution}
    \vec{z}_t = \Gamma \vec{z}_{t - 1} + w_t, \;\;\; w_t \sim \mathcal{N}_{k}(\gamma_0 e_1, W)
\end{align}
\begin{align}
\label{lds_measurement_model}
    y_{t} = C \vec{z}_t + v_t, \;\;\; v_t \sim \mathcal{N}(0, \sigma_r^2)
\end{align}
% \ziping{This is not true. You have to add some additional noise to make the noise variance for all arms to be the same. The correct statement should be you define a new variables say $\tilde{y}_t = y_t + \text{your noise}$.}
% \alt{added dummy noise variable $x_t$ to make the noise variance for all actions equal to $\sigma_r^2$}
\begin{align}
\label{lds_reward}
    r_t(a) = c_a^\top \vec{z}_t + \mu_a + \epsilon_t(a), \;\;\; \epsilon_t(a) \sim \mathcal{N}(0, \beta_a^2 \sigma_r^2) \;\;\; %\forall a \in \mathcal{A}
\end{align}
where
\begin{align*}
\vec{z}_{t} := \begin{bmatrix}
        z_{t} & z_{t - 1} & \cdots & z_{t - k + 1}
    \end{bmatrix}^\top \in \mathbb{R}^{k}
\end{align*}
See Appendix~\ref{linear_dynamic_system_proof} for exact forms for $\Gamma, W, C, c_a$.
\end{lemma}

\begin{proof}
    See Appendix~\ref{linear_dynamic_system_proof}.
\end{proof}

Lemma~\ref{linear_dynamic_system} shows that we can rewrite the process as a linear dynamical system with Gaussian noise with a specific form for the measurement model. 
Since this LDS is in companion form \citep{bellman1970structural}, the LDS satisfies structural identifiability \citep{bellman1970structural}, and therefore is observable (Assumption~\ref{system_is_observable}).

% \asim{I am realizing that the core argument we need to make above is less about whether this reduction is practically more favorable from an algorithmic perspective (i.e., which will work better methodologically) and more about theoretical analysis - namely, the systems and control literature is more concerned about stability and boundedness analyses (i.e., proving that the controller will help the system be well-behaved, or stable, and that estimation errors for parameters will be bounded and/or converge to zero similar to observability argument) and less concerned about analyses of regret. I tried to edit the paragraph accordingly to emphasize that our reduction is more favorable from an online learning theory and regret analysis perspective given the well-developed theory in the area for linear contextual bandits.}
% \alt{thank you!}

A natural approach to predicting the mean reward for each action is first to predict $\vec{z}_t$.
Since Assumption~\ref{system_is_observable} holds, one may be motivated to use the steady-state Kalman filter (Appendix~\ref{steady_state_kalman_filter}) to infer $\vec{z}_t$.
% leveraging well-developed theory for linear contextual bandits.
% \alt{usually in system settings, they assume you have a nice data set to learn the system parameters and you treat them as the true parameters. We would have to be stuck doing EM online and EM online may be computationally expensive.}
Given the ground-truth system parameters $\Gamma, C, \gamma_0, W, \sigma_r^2$, the Kalman filter prediction $\Tilde{z}_t$ is the optimal (least mean square) estimate for latent state $\vec{z}_t$ \citep{Kailath2000}.
However, we do not assume agents have access to ground-truth parameters or quality batch data for learning them offline \citep{Ljung1999}. 
% Well-established system identification approaches exist to learn system parameters offline \citep{Ljung1999}, but we also do not assume access to a quality batch data set for offline learning.
% Recent work on online system identification in the adaptive control literature 
% learn system parameters online while inferring latent states \citep{Annaswamy2021, Subbarao2016}, but they focus primarily on stability for control and boundedness or convergence of parameter estimation errors, rather than regret analysis.
While one can learn system parameters online \citep{Annaswamy2021, Subbarao2016}, we show in the following sections that it is not required to explicitly learn system parameters for forming good mean reward predictions.
% we argue that reducing the problem to a linear contextual bandit is more desirable
% However, EM is computationally expensive, difficult to implement, and difficult to analyze theoretically. 
%% we don't need to know the order of the system (i.e., k)%%
% Online system identification  \citep{Annaswamy2021, Subbarao2016} requires knowledge of the dynamic model such as the dimension of $\Gamma$.
% Asim says: you could set the model order to be a hyperparameter that you select prior to learning, which is exactly the same thing we do with our s parameter
Instead, the reduction to a linear contextual bandit allows us to implicitly learn system parameters by learning a single reward parameter (Lemma~\ref{reward_w_tilde_z}) and to leverage the well-established theory on linear bandits for analyzing regret. 

% One will see later that the algorithm we developed (Section~\ref{sec_alg}) does not require quality knowledge of many of the system parameters such as $\mu_z, W, \mu_y, V$, and the AR process's order, $k$.
% We instead reduce our problem to a linear contextual bandit and see later that this reduction enables the algorithm in Section~\ref{sec_alg} to not require knowledge of the AR process's order, $k$, nor the noise variances. 
% This is analogous to model free control (https://arxiv.org/abs/1305.7085), where control actions are selected without explicit knowledge of the underlying dynamical system parameters - only that the underlying system obeys certain dynamics. Importantly, online system identification techniques typically assume a priori knowledge of the dynamic model or autoregressive process's model order due to their focus on learning the dynamical system's parameters $\Gamma, C, \mu_z, W, \mu_y, V$ \citep{Chiuso2019, Yuen2015}. 
\subsection{Reduction to a Linear Contextual Bandit}
\label{sec_reduction}
%%% motivate why we are reducing to a linear contextual bandit and not learning parameters using MLE based methods such as EM. We do not want to do these methods because we want to run this algorithm in a real trial and want something really stable. %%%
% Asim Notes: In fact, we never actually learn the system parameters explicitly. We are able to circumvent learning those parameters in a "model free" sort of way, where "model free" in the control literature means that you control the system without explicitly learning dynamical system parameters. If you can circumvent the need to learn additional parameters and specify things like model order and such because all of that is unknown, that is an advantage of this method over the mature online system ID approaches out there, even if those methods are robust

To re-frame the problem as a linear contextual bandit, we use the converted LDS (Lemma~\ref{linear_dynamic_system}) to show that the reward (Equation~\ref{linear_reward}) can be re-written in a linear form of past history
and the steady-state Kalman filter prediction of state. This is a modified version of the decomposition in \citet{gornet2022stochastic}. 

% \ziping{Throughout the paper, we do not take advantage of the structure of $\theta_a$. I think when you present Lemma 3.4., you may simply say "this holds for some $\theta_a$".} \alt{fixed!}

\begin{lemma}
\label{reward_w_tilde_z}
    (Linear Contextual Bandit Reduction)
    Let $\mathcal{H}_{t - 1} := \sigma(a_1, r_1,...,a_{t - 1}, r_{t - 1})$ be the current history observed up to time $t$, $\Tilde{z}_t := z_{t | t - 1} = \mathbb{E}[\vec{z}_t | \mathcal{H}_{t - 1}]$ be the steady-state Kalman filter for $\vec{z}_{t}$ and let $\Tilde{r}_t(a) = \mathbb{E}[r_t(a) | \mathcal{H}_{t - 1}] = c_a^\top \Tilde{z}_t + \mu_a$.
    For a choice of $s > 0$ that controls the number of past time steps to include in the context, there exists some $\theta_a \in \mathbb{R}^{2s\cdot|\mathcal{A}| + 1}$ where the reward (Equation~\ref{linear_reward}) for action $a$ is:

\begin{align}
\label{eqn_reward_wrt_tilde_z}
    r_t(a) = \Phi_t(s)^\top \theta_a + b_t(a, s) + \varepsilon_{a; t} 
\end{align}
% \begin{align}
% \label{eqn_reward_wrt_tilde_z}
%     r_t(a) = \Phi(R_t, A_t)^\top \theta_a + \langle c_a, (\Gamma - \Gamma K C)^s \Tilde{z}_{t - s} \rangle + \varepsilon_{a; t} 
% \end{align}
where
% \ziping{for consistency, you might want to use $\Phi_t(s)$ instead of $\Phi(s)_t$.} \alt{Done!}
\begin{align}
\label{current_context}
    \Phi_t(s) := \Phi(R_t, A_t) = \begin{bmatrix}
        R_t & A_t & 1
    \end{bmatrix}^\top \in \mathbb{R}^{2s\cdot|\mathcal{A}| + 1}
\end{align}
\begin{align}
\label{eqn_bias}
    b_t(a, s) := \langle c_a, (\Gamma - \Gamma K C)^s \Tilde{z}_{t - s} \rangle
\end{align}
\begin{align}
    \varepsilon_{a; t} := r_t(a) - \Tilde{r}_t(a) = \langle c_a, \vec{z}_t - \Tilde{z}_t \rangle + \epsilon_t(a)
    \sim \mathcal{N}(0, c_a^\top P c_a + \beta_a^2\sigma_r^2)
\end{align}
\end{lemma}
% \begin{align}
%     \theta_a = \begin{bmatrix}
%         \Tilde{\beta}_a  & \Tilde{\mu}_a  & \mu_a
%     \end{bmatrix}^\top \in \mathbb{R}^{2s\cdot|\mathcal{A}| + 1}
% \end{align}
\begin{align*}
    R_t := \begin{bmatrix}
        r_{t - s} e_{a_{t - s}}^\top & \cdots & r_{t - 1} e_{a_{t - 1}}^\top
    \end{bmatrix}^\top \in \mathbb{R}^{s \cdot |\mathcal{A}|}
\end{align*}
\begin{align*}
    A_t := \begin{bmatrix}
        e_{a_{t - s}}^\top & \cdots & e_{a_{t - 1}}^\top
    \end{bmatrix}^\top \in \mathbb{R}^{s \cdot |\mathcal{A}|}
\end{align*}
%%%%%%%%%
% \begin{align*}
%     \Tilde{\beta}_a := \begin{bmatrix}
%         \frac{g_a^{s - 1}}{\beta_{1}} & \cdots & \frac{g_a^{s - 1}}{\beta_{|\mathcal{A}|}} & \cdots & \frac{g_a^{0}}{\beta_{1}} & \cdots & \frac{g_a^{0}}{\beta_{|\mathcal{A}|}}
%     \end{bmatrix}^\top \in \mathbb{R}^{s \cdot |\mathcal{A}|}
% \end{align*}
% \begin{align*}
% \Tilde{\mu}_a := \begin{bmatrix}
%         \frac{\mu_1 g_a^{s - 1}}{\beta_1} & \cdots & \frac{\mu_{|\mathcal{A}|} g_a^{s - 1}}{\beta_{|\mathcal{A}|}} & \cdots & \frac{\mu_1 g_a^{0}}{\beta_{1}} & \cdots & \frac{\mu_{|\mathcal{A}|} g_a^{0}}{\beta_{|\mathcal{A}|}}
%     \end{bmatrix}^\top \in \mathbb{R}^{s \cdot |\mathcal{A}|}
% \end{align*}
% \begin{align*}
%     g_a^j := c_a^\top (\Gamma - \Gamma K C)^j \Gamma K
% \end{align*}
%%%%%%

\begin{proof}
    See Appendix~\ref{reward_w_tilde_z_proof}.
\end{proof}

% \ziping{Need to think about how to present this. Some details might not be relevant at the moment.}
% \alt{I took a stab at it. Let me know what you think.}

Equation~(\ref{eqn_reward_wrt_tilde_z}) shows a standard linear contextual bandit problem with an additional bias term $b_t(a, s)$. $\Phi_t(s)$ is the current context obtained from a feature mapping of the $s$ most recent previous actions and rewards; $\theta_a$ incorporates the underlying LDS parameters and is the parameter to learn.
Notice that $\varepsilon_{a; t}$ is independent of history and therefore has mean 0 conditioned on history $\mathcal{H}_{t - 1}$. 
% \alt{Anna Check: Susan says bias term is is because of noise process in latent state and we never observe the latent state. NO. bias term is incurred if we do not have ground-truth parameters and if we do not set s dynamtically equal to t. I have ran simulations with our algorithm that does so and performance is identical to Kalman filter with ground-truth parameters.}

% \alt{We want to justify that our prediction of the reward is reasonable: 
% (1) Given the true underlying parameters, the Kalman filter is optimal.
% (2) If one had the true $\theta_a$ which includes the underlying parameters, then choosing $s = t$ is the same as the Kalman filter prediction.
% }
% Lemma~\ref{reward_w_tilde_z} justifies that the prediction of the reward maintained by Algorithm~\ref{alg_latent_ar_ucb}, presented in the following section, is reasonable. 
Lemma ~\ref{reward_w_tilde_z} justifies the rationale of solving the non-stationary bandit problem through a contextual bandit algorithm, say LinUCB.
By selecting $s$, one is selecting the number of recent time steps used to predict the current mean reward.
If one had access to ground-truth $\theta_a$ and dynamically sets $s = t$, then such an agent's performance is the same as an agent that predicts the mean reward using a steady-state Kalman filter (Appendix~\ref{steady_state_kalman_filter}) with ground-truth parameters. Recall that with the true underlying parameters, the Kalman filter estimate $\Tilde{z}_t$ is the optimal estimate for latent state $z_t$.
However, because we do not assume access to ground-truth parameters and must learn parameters online, one must set $s$ to balance bias and variance, which allows for a good approximation of the Kalman filter estimate. This bias-variance trade-off,  controlled by $s$, also appears in the regret bound (Theorem~\ref{thm_regret}) presented later.
% If an agent selects $s = 0$, then the agent has the same action-selection and update strategy as a standard, stationary MAB.
This linear bandit reduction is also desirable because an agent only needs to specify a value for $s$ and does not need to know the ground-truth AR order $k$, the initial state $\vec{z}_0$, nor the noise variances in practice for forming an estimator for $\theta_a$.

%%%%%%%%% OLD STUFF %%%%%%%%%%
% \subsubsection{Linear Bandit with Inaccurate Context}
% If one observes $Z_t$, then Equation~\ref{measurement_model} becomes the reward for a standard linear bandit: $Z_t$ is the true context and $(C_a)_{a \in \mathcal{A}}$ are the unknown reward parameters. However, since $Z_t$ is latent, the agent can only form a prediction $\hat{Z}_t$ of $Z_t$. Thus, the problem setting reduces to a linear bandit with inaccurate context as one can interpret $\hat{Z}_t$ as the inaccurate context of the true underlying context $Z_t$.

% We now redefine the problem statement. At time step $t$, the agent forms a predicted context $\hat{Z}_t$. Since the agent never observes the true context $Z_t$, the history up to and including time step $t$ is $\mathcal{H}_t := \{(\hat{Z}_\tau, a_\tau, r_\tau)\}_{\tau = 1}^t$. Let $\delta_t = \hat{Z}_t - Z_t$ be the context error. Notice that because of the noise in the state process $\xi_t$ and the reward noise $\epsilon_t(a)$, the context error is non-vanishing. The agent's goal is to, at every time step $t$: (1) design a predictor $\hat{Z}_t = f_{\theta}(\mathcal{H}_{t - 1})$ and (2) design a policy $\pi_t(\cdot | \mathcal{H}_{t - 1}, \hat{Z}_t)$ using that predictor as to maximize the reward by taking action $a_t \sim \pi_t(\cdot | \mathcal{H}_{t - 1}, \hat{Z}_t)$.

\subsection{Assumptions}
We make the following regularity assumptions on the environment.

% \alt{Assumption~\ref{assump_state_stability} is needed for the bias term $b_t(a, s)$ in Equation~\ref{eqn_reward_wrt_tilde_z} to go away with large $s$.}

\begin{assumption}
\label{assump_ar_process}
    (Stability and Boundedness of the AR Process) For the parameters $\gamma_0, \gamma_1,...,\gamma_k$ of the latent AR process, 
    $| \sum_{j = 1}^k \gamma_j| < 1$ and $|\gamma_0| \leq c$ for some $c < +\infty$
\end{assumption}

% don't think we need this
% \begin{assumption}
% \label{bounded_obs_covariance}
%     (Bounded Observation Covariance) There exists some matrix $R$ such that $R \succ R_{t}$ for all $t \in [T]$. $A - B \succ 0$ means $A-B$ is positive definite
% \end{assumption}
% \begin{assumption}
% \label{reward_bounded}
%     (Boundedness of Reward) With probability at least $1 - \delta_r$ $r_t \leq C_r$ for all $t$
% \end{assumption}
% \begin{assumption}
% \label{action_set_assumption}
%     (Boundedness of Context) $\|\Phi(s)_t\| \leq L(s)$ for all $t$
% \end{assumption}

\begin{assumption}
\label{reward_param_assumption}
    (Bounded Reward Parameter) For every action $a \in \mathcal{A}$, $\|\theta_a\| \leq S_a$ for $S_a \in \mathbb{R}^+$
\end{assumption}

Assumption~\ref{assump_ar_process} is standard for AR processes. Assumption~\ref{reward_param_assumption} is standard for theoretical results for linear bandits \citep{abbasi2011improved}.
More importantly, Assumption~\ref{assump_ar_process} implies the stability of state transition matrix $\Gamma$, a common assumption for LDS \citep{bertsekas2012dynamic}. Namely, the parameters $\gamma_1,...,\gamma_k$ of the latent AR process form $\Gamma$ such that $|\lambda_{\max}(\Gamma)| < 1$.
The stability of $\Gamma$ is needed for the bias term $b_t(a, s)$ in Equation~\ref{eqn_reward_wrt_tilde_z} to decrease with large $s$ as $|\lambda_{\max}(\Gamma)| < 1 \implies |\lambda_{\max}(\Gamma - \Gamma K C)| < 1$ \citep{anderson2005optimal}.

% \begin{assumption}
% \label{assump_state_stability}
% (Stability of State Transition Matrix) % https://arxiv.org/pdf/1807.09120
% The parameters $\gamma_1,...,\gamma_k$ of the latent AR process form state process matrix $\Gamma$ such that $|\lambda_{\max}(\Gamma)| < 1$.
% \end{assumption}
\subsection{Regret}
\label{sec_regret_def}
We define regret with respect to the dynamic oracle \citep{besbes2014stochastic}, the standard choice in the non-stationary bandit settings. The dynamic oracle observes all information in the environment (including $z_t$) and then acts optimally with that information. The oracle therefore knows the true reward means $\mathbb{E}[r_t(a) | \vec{z}_t]$ at every time step $t$ for each action $a$ and selects the optimal action for every $t$: $a_t^* = \arg \max_{a \in \mathcal{A}} \mathbb{E}[r_t(a) | \vec{z}_t]$. The regret with respect to the dynamic oracle is:
% \alt{Note: we are defining a random regret right now and offering a high-probability bound. So regret is a r.v.}
\begin{align}
    \label{eqn_regret}
    \text{Regret}(T; \pi) =
    \sum_{t = 1}^T  \mathbb{E}[r_t(a_t^*) - r_t(a_t) | \vec{z}_t]
\end{align}
where $a_t$ is the action selected by the algorithm at time step $t$ following policy $\pi$. 

Achieving sub-linear regret against the dynamic oracle is not guaranteed without assuming vanishing non-stationarity in the environment.
% \alt{We need to clearly mention how hard this problem is and why, dependent on $\sigma_z^2$, it's not guaranteed to achieve a sub-linear regret bound. Instead, our goal is to achieve a high probability regret bound that is interpretable and helpful but be clear that for most environments, it's linear in $T$.}
Since our latent AR setting does not make this assumption, our goal is to provide an interpretable regret bound with respect to the non-stationarity in the environment. 
For a full discussion of the regret definition and the difficulty in achieving sub-linear regret, see Appendix~\ref{app_regret_disc}.

% \paragraph{Dynamic Regret} Dynamic regret \citep{besbes2014stochastic} is defined with respect to a dynamic oracle that follows the optimal dynamic sequence of actions. Namely, the dynamic oracle knows the optimal action at every time step $t$ and therefore achieves the largest expected reward at every time step $t$. This is akin to the ``standard oracle" defined above that observes all information in the environment and then acts optimally with that information. \citet{besbes2014stochastic} was able to achieve sub-linear regret because their setting assumes a finite constant (variation budget) of how much the mean rewards can change over time. Their regret bound is with respect to this constant. In our setting, such an oracle would be too strong as we do not make any assumptions on the total variation of the expected rewards over $T$ time steps.

\section{LinUCB Algorithm for Latent AR Bandits}
\label{sec_alg}
%%% GIVE A HIGH-LEVEL SUMMARY OF THE ALGORITHM %%%
% Note: we do not assume to know the parameters of the LDS. We implicitly learn system parameters along with reward parameters in one big vector theta_a
We present our algorithm coined Latent AR Bandit LinUCB (LARL), shown in Algorithm~\ref{alg_latent_ar_ucb}. LARL is based on the LinUCB algorithm \citep{li2010contextual, abbasi2011improved} for linear contextual bandits, modified to handle our non-stationary environment. 
For a fixed choice of $s > 0$, LARL uses rewards and actions from the $s$ most recent time steps to form current context $\Phi_t(s)$ (Equation~\ref{current_context}).
By carefully constructing the context $\Phi_t(s)$ this way, we can implicitly predict the latent state $z_t$ and efficiently learn the parameters $\theta_a$ online.
% and handle the un-observability of $z_t$ (i.e., the cause of the non-stationarity).

As a review, LinUCB maintains regularized least squares (RLS) estimators for each action $a$ at time step $t$:

\begin{align}
\label{eqn_rls_estimator}
    \hat{\theta}_{a, t} = V_{a ,t}^{-1} b_{a, t}
\end{align}
where
$V_{a ,t} = \lambda I + \sum_{j = 1}^{t} \mathbb{I}[a_j = a] \Phi_j(s) \Phi_j(s)^{\top}$, and $b_{a, t} = \sum_{j = 1}^{t} \mathbb{I}[a_j = a] r_j \Phi_j(s)$

For action-selection, LinUCB forms a confidence set $\mathcal{C}_{a, t - 1}$ using the most recent RLS estimator $\hat{\theta}_{a, t - 1}$ for every action $a$ and selects the action with the highest confidence bound on its reward:

\begin{align}
    \label{action_selection_eqn}
      a_t = \underset{a \in \mathcal{A}}{\arg \max} \;\; \underset{\theta_a \in \mathcal{C}_{a, t - 1}}{\max} \Phi_t(s)^T \theta_{a}
\end{align}

%% ALTERNATIVE %%
% \begin{align}
%     \label{action_selection_eqn}
%     a_t = \underset{a \in \mathcal{A}}{\arg \max} \;\; \Phi_t(s)^T \hat{\theta}_{a, t - 1} + \beta_{a, t - 1}(\delta) \|\Phi_t(s)\|_{\hat{V}_{t - 1}^{-1}}
% \end{align}

Notably, our algorithm requires no knowledge of the ground-truth AR order $k$, initial state $\vec{z}_0$, nor noise variances of the state and reward processes.
Notice also that for $s = 0$, Algorithm~\ref{alg_latent_ar_ucb} reduces to the standard UCB method for stationary MABs.

To show theoretical results for LARL, we first show that for each action $a$, $\theta_a$ lies in the confidence set $\mathcal{C}_{a, t}$ for all $t$ (Lemma~\ref{lemma_ellipsoid}). The radius of this confidence set is an enlarged version of the one presented in \cite{abbasi2011improved} to account for the bias term.
Finally, we use Lemma~\ref{lemma_ellipsoid} to prove the regret bound in Theorem~\ref{thm_regret}.

%%%%% ALGORITHM %%%%%
\begin{algorithm}[t]
\caption{Latent AR LinUCB}
% \DontPrintSemicolon
\begin{algorithmic}[1]
\label{alg_latent_ar_ucb}
  \STATE {\bfseries Inputs:} 
  $V_{a, 0} = \lambda I, b_{a, 0} = \vec{0}$, $\theta_{a, 0} = \vec{0}$ for all $a \in \mathcal{A}$, $s \in \mathbb{N}$
  \FOR{$t = 1, 2, ..., T$}
  \STATE Use rewards and actions from the most recent $s$ time steps $r_{t - s},...,r_{t - 1}, a_{t - s}, ..., a_{t - 1}$ to form current context $\Phi_t(s)$ defined in Equation~\ref{current_context}.
  \STATE For each action $a$, use most recent RLS estimator $\hat{\theta}_{a, t - 1}$ to form confidence set $\mathcal{C}_{a, t - 1}$.
  \STATE Select action $a_t$: 
  $$a_t = \underset{a \in \mathcal{A}}{\arg \max} \;\; \underset{\theta_a \in \mathcal{C}_{a, t - 1}}{\max} \Phi_t(s)^T \theta_{a}$$
  % ALTERNATIVE %
  % \STATE Select action $a_t$: 
  % $$a_t = \underset{a \in \mathcal{A}}{\arg \max} \;\; \Phi_t(s)^T \hat{\theta}_{a, t - 1} + \beta_{a, t - 1}(\delta) \|\Phi_t(s)\|_{\hat{V}_{t - 1}^{-1}}$$
  \STATE Execute action $a_t$ and observe $r_t$.
  % setting noise estimate
  \STATE Update history $\mathcal{H}_t = \{(\Phi(s)_{t'}, a_{t'}, r_{t'})\}_{t' = 1}^t$
  \STATE Update RLS estimator for action $a_t$ as in Equation~\ref{eqn_rls_estimator}: 
  \STATE
    $\hat{\theta}_{a_t, t} = V_{a_t ,t}^{-1} b_{a_t, t}$,
   \STATE $V_{a ,t} = V_{a, 0} + \sum_{j = 1}^{t} \mathbb{I}[a_j = a] \Phi_j(s) \Phi_j(s)^{\top}$, \; \\ $b_{a, t} = b_{a, 0} + \sum_{j = 1}^{t} \mathbb{I}[a_j = a] r_j \Phi_j(s)$
  \ENDFOR
\end{algorithmic}
\end{algorithm}

\subsection{Confidence Set}
\begin{lemma}
\label{lemma_ellipsoid}
    [Confidence Set for Latent AR Bandits] Suppose Assumptions \ref{assump_ar_process} and \ref{reward_param_assumption} holds. For given action $a$, with probability at least $1 - \delta$
    % $1 - \delta_\beta - \delta_r$ \alt{Is this prob. correct as well?}, 
    where $\delta \in (0, 1)$, the true parameter $\theta_a$ (Definition~\ref{eqn_reward_wrt_tilde_z}) is in the confidence ellipsoid $\mathcal{C}_{a, t}$ centered at $\hat{\theta}_{a, t}$ (Equation~\ref{eqn_rls_estimator}), for all $t \in [T]$:
    \begin{align}
    \label{our_confidence_set}
       \mathcal{C}_{a, t} := \{\theta_a \in \mathbb{R}^d \; | \; \| \hat{\theta}_{a, t} - \theta_a\|_{V_{a, t}} \leq \beta_{a, t}(\delta)\} 
    \end{align}
    where
    \begin{multline}
    \label{eqn_my_radius}
        \beta_{a, t}(\delta) = R \sqrt{(2s|\mathcal{A}| + 1) \log \bigg(\frac{1 + n_{a, t} L(s, \delta/2) / \lambda}{\delta/2}\bigg)} \\
        + \sqrt{\lambda} S_a 
        + \tau(a, s)_t \sqrt{\sum_{j = 1}^t \mathbb{I}[a_j  = a] b_j(a, s)^2} 
    \end{multline} 
    where $n_{a, t} = \sum_{j = 1}^t \mathbb{I}[a_j = a]$ and
    \begin{align}
    \label{eqn_tau}
        \tau(a, s)_t = \sqrt{\sum_{j = 1}^t \mathbb{I}[a_j = a] \| \Phi_j(s) \|^2_{V_{a, t}^{-1}}}
    \end{align}
\end{lemma}
\begin{proof}
    See Appendix~\ref{confidence_ellipsoid_proof}
\end{proof}

% Now based on Lemma~\ref{lemma_ellipsoid_known_k}, an algorithm can select actions as follow:
% \begin{align}
%     \label{action_selection_eqn}
%     a_t = \underset{a \in \mathcal{A}}{\arg \max} \;\; \Phi(s)_t^T \hat{\theta}_{a, t - 1} + \beta_{a, t - 1}(\delta) \|\Phi(s)_t\|_{\hat{V}_{t - 1}^{-1}}
% \end{align}

% An algorithm that has such an action-selection procedure obtains the regret upper bound in Theorem~\ref{thm_reg_known_k} found below.
\subsection{Regret Bound}
We now derive a regret bound for LARL (Algorithm~\ref{alg_latent_ar_ucb}).
The main proof idea is to introduce an intermediate agent that knows the ground-truth parameters and uses the steady-state Kalman filter prediction to select actions. We are able to add and subtract the mean reward obtained by such an agent to the instantaneous regret. The instantaneous regret therefore decomposes into the regret of the dynamic oracle against the intermediate agent and the regret of the intermediate agent against LARL.
\begin{theorem}
\label{thm_regret}
    Suppose all the assumptions mentioned in Lemma~\ref{lemma_ellipsoid} hold.
    % and Assumption~\ref{assump_state_stability} hold. 
    With probability at least $1 - \delta$
    % $1 - \delta_\beta - \delta_r - \delta_z$ \alt{is this prob. correct? there's a $\delta_r$ to tail bound the context, $\delta_\beta$ in the confidence set and a $\delta_z$ in the regret bound?}, 
    where $\delta \in (0, 1)$, the regret of Algorithm~\ref{alg_latent_ar_ucb} in the non-stationary latent AR bandit environment (Definition~\ref{def_non_stat_latent_auto_bandit}) is bounded as follows:
\begin{gather*}
    \text{Regret}(T; \pi_{\text{LARL}})
    \leq 8 \max_a \|c_a\| \sqrt{\frac{\sigma_z^2}{1 - \sigma_{\max}(\Gamma)^2}}
    \sqrt{2(k + \log(3 / \delta))} \cdot T \\
    + 2 \beta_{T}(2\delta/3) \sqrt{\sum_{t = 1}^T \|\Phi_t(s)\|_{V_{a_t, t - 1}^{-1}}^2} \sqrt{T} + 2 \sum_{t = 1}^T \max_a |b_t(a, s)|
\end{gather*}
    for $\beta_T(\delta') = \max_a \beta_{a, T - 1}(\delta')$ (Equation~\ref{eqn_my_radius})
    % }$\tau_T$ (Equation~\ref{eqn_tau}).
\end{theorem}

% \ziping{you may present the special case where $\delta_\beta = \delta_r = \delta_z = \delta/3$. So the total failure prob is $1-\delta$.}
% \alt{thanks for the suggestion, I made the change to Lemma~\ref{lemma_ellipsoid} and Theorem~\ref{thm_regret}. Could you double check that I wrote it correctly?} \alt{also do we have to handle cases when the probability of events is dependent?}
\begin{proof}
    See Appendix~\ref{regret_proof}
\end{proof}

% \ziping{Should mention that the first term represents the gap between the optimal decision making based on Kalman filtering prediction $\tilde{z}_t$ and the dynamic oracle.\\
% The second term represents the complexity of learning to behave like the optimal policy based on Kalman filtering. The third term represents the bias. \\
% Then mention that $s$ balances bias and variance. 
% } \alt{done!}

The first term of the regret represents the gap between the optimal decision-making based on the Kalman filter prediction $\tilde{z}_t$ and the dynamic oracle. This term also captures the non-stationarity of the environment as controlled by $\sigma_z^2$, the noise variance of the latent AR process. The second term represents the complexity of learning to behave like the optimal policy based on Kalman filtering.
% and is the standard result for the LinUCB algorithm in stationary environments. 
The third term captures the bias in the reward function. 

The first term captures how the regret rate is ultimately dependent on $\sigma_z^2$. For a fixed $T$ in environments where $\sigma_{\max}(\Gamma) \leq 1 - \epsilon$ for $\epsilon > 0$ and $\sigma_z^2 = T^{c - 2}$ for some constant $c < 2$, our algorithm achieves sub-linear regret.
For example, if $\sigma_z^2 = \frac{1}{T}$, then the regret is on the order of $\sqrt{T}$.
This is congruent with a variety of other non-stationary bandit formulations where the non-stationarity budget is sub-linear in $T$. See Appendix~\ref{sublinear_regret_possible} for more details.
However, if $c \geq 2$, then $\sigma_z^2$ is too large and the regret is not sub-linear. This is because for large $\sigma_z^2$, even with ground-truth parameters, the most optimal prediction $\Tilde{z}_t$ used for predicting $\Tilde{r}_t(a)$, can be far away from the realization of $z_t$ and therefore $r_t$.

The second and third terms describe the bias-variance trade-off the algorithm designer makes with the choice of $s$. For large $s$ the bias decreases, as $s \rightarrow \infty \implies$ $b_t(a, s) \rightarrow 0$ for any $a$, however, the dimensionality of $\Phi_t(s)$ increases (Lemma~\ref{bounded_context_lemma}) which increases the variance. See Appendix~\ref{bias_var_exp} for simulations that verify this trade-off empirically.

% \ziping{Need to compare to some results in the literature that assumes bounded non-stationarity. Their results should also have a linear in T term. I am assuming that term will be loose when $c \geq 2$.}
% \alt{we compare to other results in Appendix.}
\section{Experiments}
\label{sec_experiemnts}
Through simulations, we highlight how our proposed algorithm LARL (Algorithm~\ref{alg_latent_ar_ucb}) can outperform various stationary and non-stationary baselines. We assess performance based on cumulative regret with respect to the dynamic oracle (Section~\ref{sec_regret_def}).
Additional experiments and results can be found in Appendix~\ref{app:additional_exps}.

For all experiments, we set $T = 200$. We consider 2 actions $\mathcal{A} = \{0, 1\}$, reward parameters $\mu_0, \mu_1 = [0,0], \beta_0, \beta_1 = [-1, 1]$, $\gamma_0 = 0$, $\sigma_z = 1$, and $\sigma_r = 1$. 
% \ziping{How do you choose $\gamma$'s?} \alt{I draw them randomly but ensure that they are "valid", namely roots are inside the unit circle}
$\gamma_1,...,\gamma_k$ are drawn randomly from a uniform distribution and post-processed to ensure Assumption~\ref{assump_ar_process} holds. 
We consider environments that vary by the AR order $k$. 
% and by the noise of the latent process, $\sigma_z$. 
%Each environment is a unique combination of $k = [1, 2, 5, 10]$ and $\sigma_z = [0.001, 1, 10]$. 
For each environment variant, we simulate $100$ Monte-Carlo trials and in each trial, the $k$ values in the initial state $\vec{z}_0$ are drawn randomly in every trial.

%%% LARL with different approaches of selecting $s$ can outperform various baselines:
We test the performance of LARL against various stationary and non-stationary baselines. The competing baselines are: (a) ``Stationary'', standard UCB which treats the environment as a stationary multi-armed bandit, (b) ``AR UCB" \citep{bacchiocchi2022autoregressive}, the UCB algorithm developed for stationary AR environments, (c) ``SW UCB'' \citep{garivier2011upper}, the Sliding Window UCB algorithm, (d) ``Rexp3'' \citep{besbes2014stochastic}, which runs the Exp3 algorithm with restarts.

To prove theoretical results, we left $s$ arbitrary. To select $s$ in simulations, we implement an ``exploration" period that selects actions randomly up to time step $t'$. Then using the data collected during the exploration period, we choose $s$ based on the Bayesian Information Criterion (BIC) and commit to that $s$ for the rest of the time-steps. For experiments, we let $t' = \lfloor T / 5 \rfloor$.
% To deal with the problem of selecting $s$, we run two variants of our algorithm: LARL-ETC and LARL-Ensemble. LARL-ETC implements an ``exploration" period which runs LARL with various values of $s$ before committing to a single $s$ for the rest of the time-steps. LARL-Ensemble runs an ensemble of LARL with various values of $s$ and uses weighted majority voting for action-selection. 
All baseline algorithms that use a UCB-based strategy (including LARL), use regularization parameter $\lambda = 1$ in each environment. 

\subsection{Results}

\begin{figure*}[t]
  \centering
  \subfigure[]{
    \includegraphics[width=0.31\textwidth]{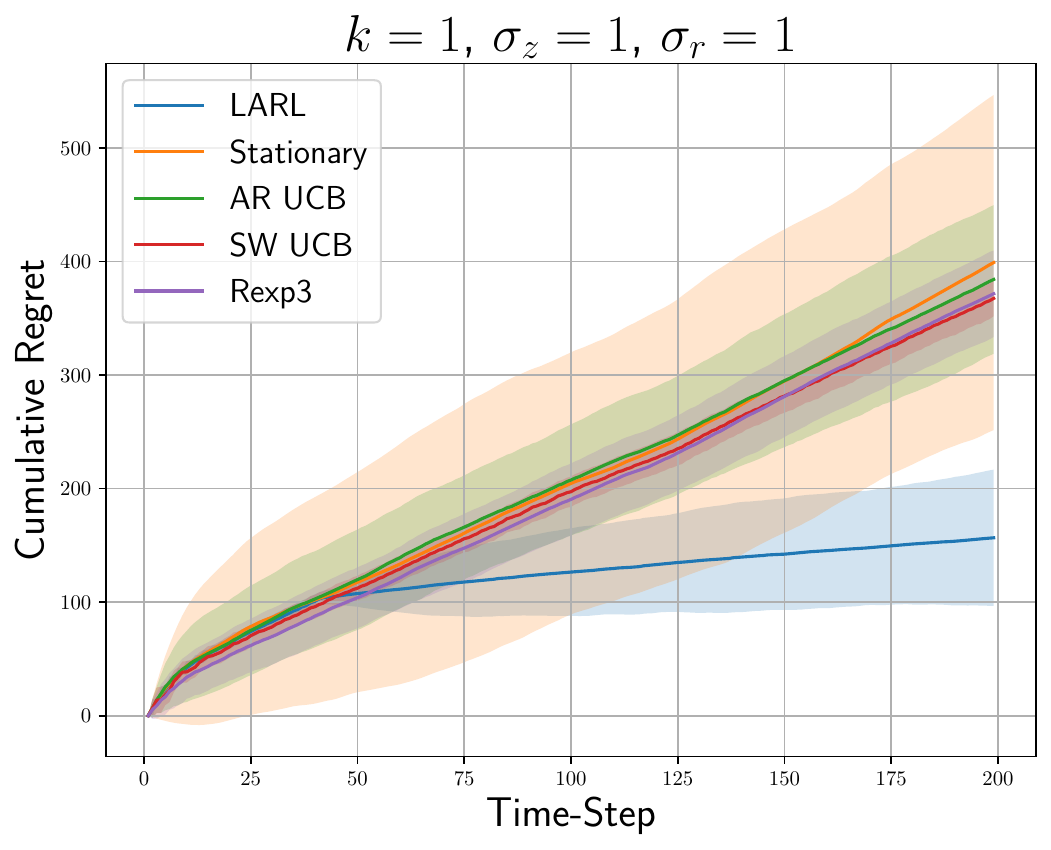}
  }
  \subfigure[]{
    \includegraphics[width=0.31\textwidth]{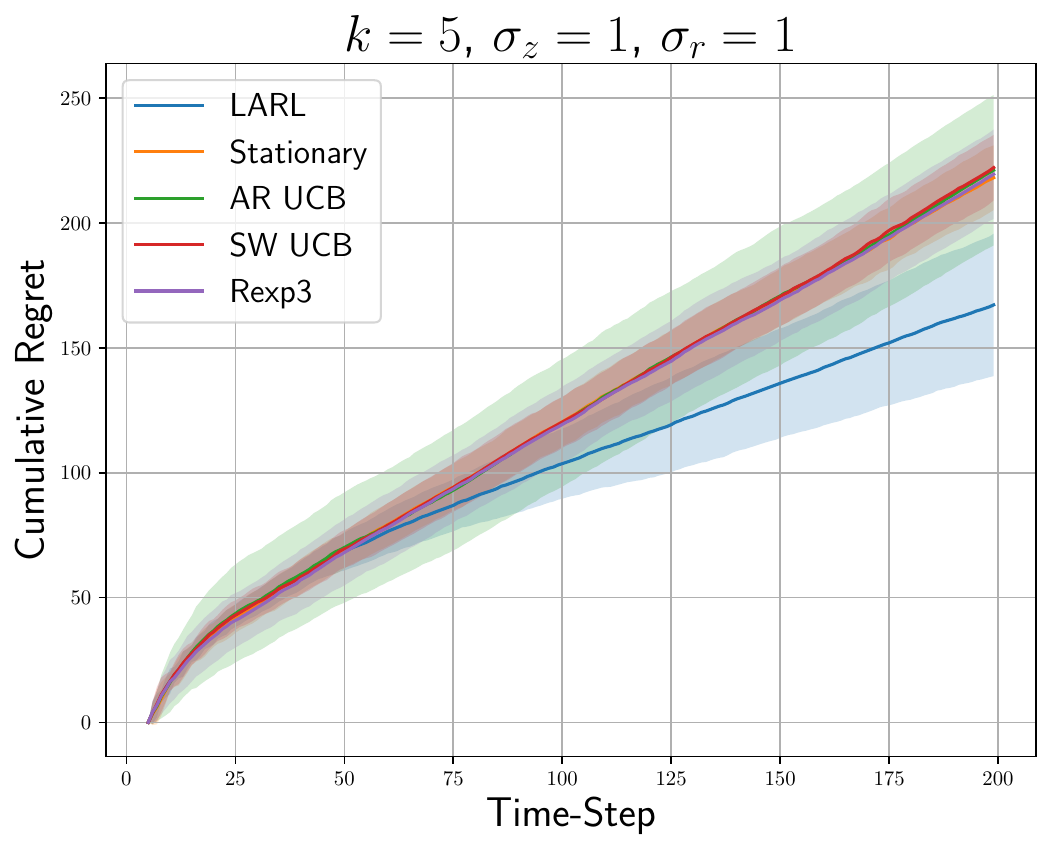}
  }
  \subfigure[]{
    \includegraphics[width=0.31\textwidth]{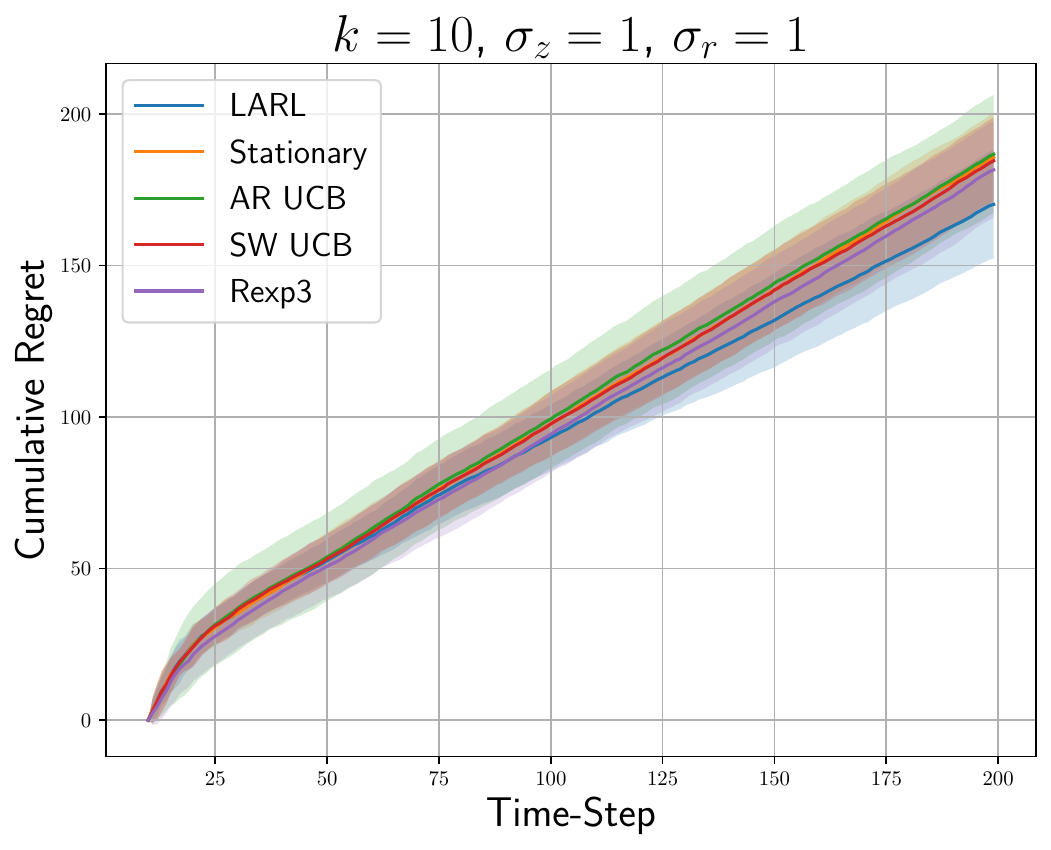}
  }
  \caption{Our algorithm LARL (blue), with $s$ chosen using BIC after a period of pure exploration, consistently achieves lower cumulative regret (Equation~\ref{eqn_regret}) over time against various baseline methods.
  Line is the average and shaded region is $\pm$ standard deviation across 100 Monte Carlo simulated trials.}
  \label{fig:main_cum_regret}
\end{figure*}

\begin{figure*}[t]
  \centering
  \subfigure[]{
    \includegraphics[width=0.45\textwidth]{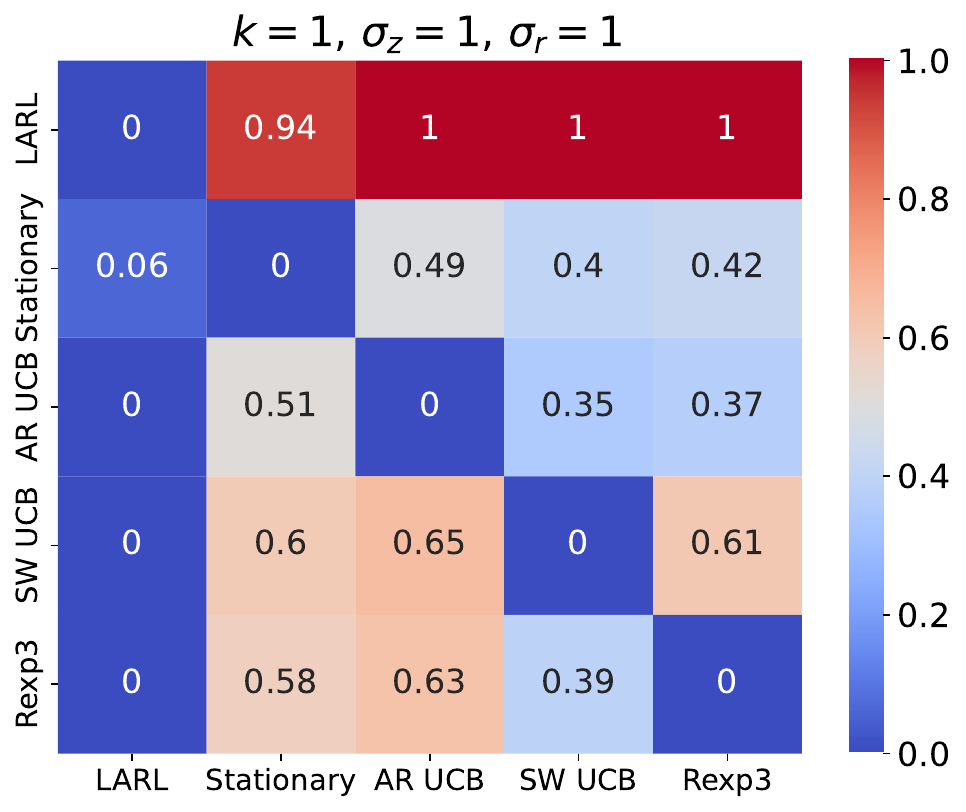}
  }
  \subfigure[]{
    \includegraphics[width=0.45\textwidth]{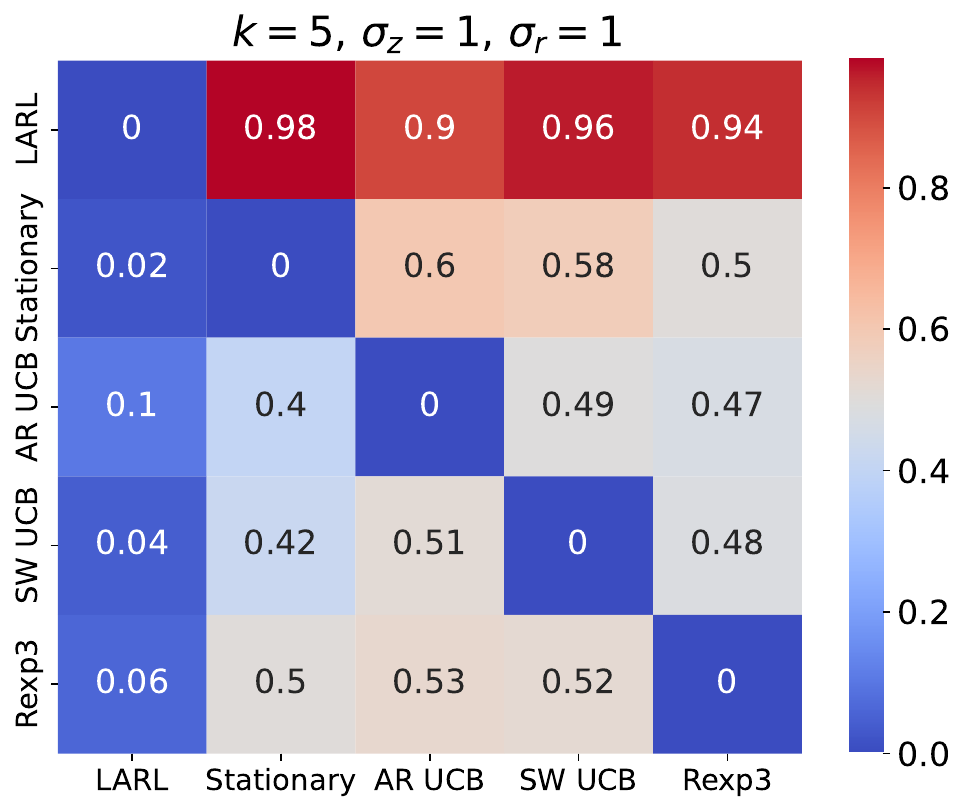}
  }
  \subfigure[]{
    \includegraphics[width=0.45\textwidth]{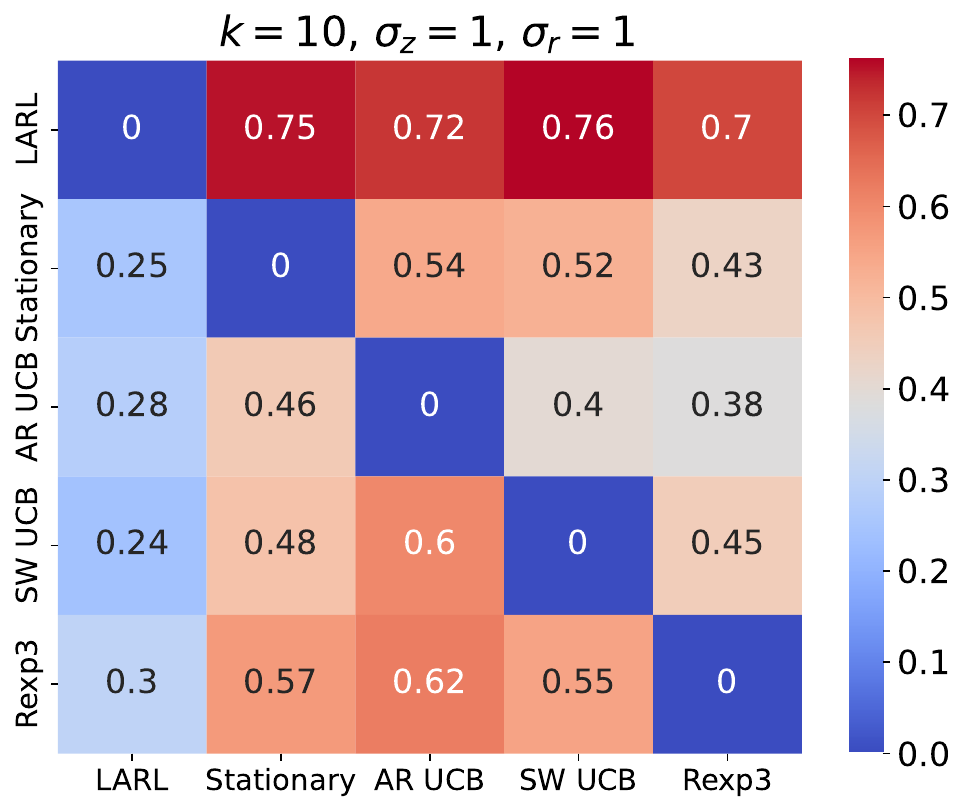}
  }
    \caption{Pairwise comparisons between algorithms in the three variants of the simulation environment where $k = 1, 5, 10$, respectively. Each cell shows the proportion of 100 Monte-Carlo repetitions where the algorithm listed in the row achieved lower cumulative regret than the algorithm listed in the column. 
    % ``Avg.'' shows the average value of each row.
    Our algorithm LARL (top row) consistently outperforms baseline methods in pairwise comparison.
    }
    \label{pairwise_comparisons_2}
\end{figure*}

%%% WHEN S IS SET REASONABLY %%%
Figure~\ref{fig:main_cum_regret} shows cumulative regret over time. Figure~\ref{pairwise_comparisons_2} shows pairwise comparisons between algorithms in terms of total cumulative regret. Our method LARL consistently outperforms baseline algorithms developed for stationary and non-stationary environments. 
% as the non-stationary environment gets more challenging.
Stationary and AR UCB were developed for stationary environments with fixed reward means and cannot adapt to the non-stationarity of the reward means.
Although developed for a non-stationary environments, SW UCB and Rexp3 perform similarly to the stationary baselines because SW UCB assumes the mean rewards remain  constant over epochs and Rexp3 assumes the non-stationarity has bounded total variation.
In the latent AR environment (Definition~\ref{def_non_stat_latent_auto_bandit}), these assumptions are not guaranteed as the non-stationary budget can be linear in $T$.
As $k$ increases, the performance of LARL approaches the performance of baseline methods because the algorithm needs to fit more parameters, and thus requires more data to learn effectively.
% because the algorithm needs to fit more parameters as $k$ increases, and thus requires more data to learn effectively.

% old notes chatting with Asim: If s is smaller relative to k (underfitting), then our algorithm is looking back less time steps than it should given the true dimensionality of the system. The smaller s is relative to k, then the larger the performance degradation because it is underestimating the length of time in the past it should look. If s is larger relative to k (overfitting), then our algorithm's reward model is over parameterized and fitting more parameters than necessary.

%%%%% possibly more experiments %%%%
% Susan says she wouldn't be surprised if our algorithm is robust to other non-stationary environments (e.g., change points)
\section{Conclusion and Future Work}
In this paper, we study the non-stationary bandit problem where the mean rewards of actions change over time due to a latent, AR process of order $k$.
We propose a new online algorithm, LARL, that leverages the structure of the non-stationary mechanism in this setting.
LARL employs the key idea that this non-stationary bandit can be reduced to a linear dynamical system and solved using a linear contextual bandit with a thoughtful design of the context space. This reduction motivates LARL's linear contextual bandit strategy to implicitly predict the current latent state $z_t$ and efficiently learn parameters online. Furthermore, with a choice of hyperparameter $s$ that trades off bias and variance, one can view the reward model of LARL as a reasonable approximation of a steady-state Kalman filter with ground-truth parameters.
% We propose a first algorithm for the latent AR setting, which incorporates past rewards and past actions into its context and does not need to know or learn the AR order $k$.
% We prove an interpretable regret bound for LARL with respect to the level of non-stationarity (controlled by $\sigma_z^2$) in the environment. LARL achieves sub-linear regret if $\sigma_z^2$ is sufficiently small with respect to $T$.
% $\sigma_z^2 = T^{c - 2}$ for some constant $c < 2$.
% Experiments show that our algorithm outperforms baseline methods in various simulation environments.

\subsection{Future Work}
A natural extension is to generalize the present work to contextual bandits with the inclusion of exogenous, observed context features. Although we have proposed an initial approach for selecting hyperparameter $s$, one can explore other approaches such as running an ensemble of agents with different $s$ or finding an optimal $s$ as a function of environment parameters $k$, $T$, $\sigma_r$, and $\sigma_z$.
% shown our algorithm empirically performs well with a reasonable choice of $s$ that trades off bias and variance. It would be helpful to have a data-adaptive procedure for selecting hyperparameter $s$.
% with respect to varying $k$ and $T$.
Lastly, to make our ideas clear, we formulated the latent state as a scalar, but one can consider a generalization of our setting where the latent state is multi-dimensional.
% and latent features are a mixture of known and unknown. 

% Finale says if a reviewer complains about multi-dimensional latent case. Then we can argue that it is natural to consider a 1d latent state that represents a score for a compilation of multiple features. Nonetheless we think this is an interesting extension for future work.

%simpler case wehre you have part of context is unrelated to the latent process (thus exogenous). 

\subsubsection*{Broader Impact Statement}
\label{sec:broaderImpact}
This paper presents work whose goal is to advance the field of reinforcement learning for use in real-world problems such as digital health. The assumptions we make in the paper may be valid for some of these domains and not others.

% Acknowledgements should only appear in the accepted version.
\section*{Acknowledgments}
This research was funded by NIH/NIDCR grant UH3DE028723, the Center for Methodologies for Adapting and Personalizing Prevention, Treatment and Recovery Services for SUD and HIV P50 DA054039, and the National Science Foundation grant no. IIS-1750358.  Any opinions, findings, and conclusions or recommendations expressed in this material are those of the author(s) and do not necessarily reflect the views of the National Science Foundation. 
SAM  holds concurrent appointments  at Harvard University and as an Amazon Scholar. This paper describes work performed at Harvard University and is not associated with Amazon.

%%%%%%%%%%%%%%%%%%%%%%%%%%%%%%%%%%%%%%%%%%%%%%%%%%%%%%%%%%%%%%%%
%% NOTE: THIS MARKS THE END OF THE "MAIN TEXT"
%%%%%%%%%%%%%%%%%%%%%%%%%%%%%%%%%%%%%%%%%%%%%%%%%%%%%%%%%%%%%%%%

%%%%%%%%%%%%%%%%%%%%%%%%%%%%%%%%%%%%%%%%%%%%%%%%%%%%%%%%%%%%%%%%
%% Bibliography
%%%%%%%%%%%%%%%%%%%%%%%%%%%%%%%%%%%%%%%%%%%%%%%%%%%%%%%%%%%%%%%%
\bibliography{main}
\bibliographystyle{rlj}

%%%%%%%%%%%%%%%%%%%%%%%%%%%%%%%%%%%%%%%%%%%%%%%%%%%%%%%%%%%%%%%%
%% Appendix
%%%%%%%%%%%%%%%%%%%%%%%%%%%%%%%%%%%%%%%%%%%%%%%%%%%%%%%%%%%%%%%%
\appendix
\section{Proofs}
\label{proofs}
\subsection{Proof of Lemma~\ref{linear_dynamic_system}}
\label{linear_dynamic_system_proof}
We first present exact forms for $W, \Gamma, C, c_a$.
% \begin{align*}
% \mu_z := \begin{bmatrix}
%         \gamma_0 & 0 & \cdots & 0
%     \end{bmatrix}^\top \in \mathbb{R}^{k}
% \end{align*}
\begin{align*}
W := \text{diag}(\sigma_z^2, 0, \cdots, 0) \in \mathbb{R}^{k \times k}
\end{align*}
% \begin{align*}
% w_{t} := \begin{bmatrix}
%         0 & \xi_{t} & 0 & \cdots & 0
%     \end{bmatrix}^\top \in \mathbb{R}^{k + 1}
% \end{align*}
% $V = \sigma_r^2 / \min_a \beta_{a}^2 \in \mathbb{R}$ is a constant that bounds the variance $\sigma_r^2 / \beta_{a}^2$ for all possible actions $a$.
\begin{align*}
    \Gamma = \begin{bmatrix}
        \gamma_1 & \gamma_{2} & \cdots & \gamma_k \\
        1 & 0 & \cdots & 0 \\
        \vdots & \vdots & \vdots & \vdots \\
        0 & \cdots &  1 & 0
    \end{bmatrix} \in \mathbb{R}^{k \times k}
\end{align*}
% \begin{align*}
%     W_t = \begin{bmatrix} 0 & \xi_t & 0 & \cdots & 0
%     \end{bmatrix}^\top \in \mathbb{R}^{k + 1}
% \end{align*}
\begin{align*}
C = \begin{bmatrix}
    1 & 0 & \cdots & 0
\end{bmatrix} \in \mathbb{R}^{1 \times k}
\end{align*}
\begin{align*}
c_a := \begin{bmatrix}
    \beta_a & 0  & \cdots & 0
\end{bmatrix}^\top \in \mathbb{R}^{k}
\end{align*}

\begin{proof}
    For the latent state evolution, it is fairly straightforward to compute that Equation~\ref{state_evolution} is equivalent to Equation~\ref{latent_state} with the definitions of $\vec{z}_t, \Gamma$, and $w_t$. Notice that the second entry in $\vec{z}_t$ corresponds to the latent state value $z_t$. Similarly, for the reward function, we can directly compute that Equation~\ref{lds_reward} is equivalent to Equation~\ref{linear_reward} with the definition of $c_a^\top$.

    %%% MEASUREMENT MODEL %%%%
    All that is left is to show the equivalence of the measurement model $y_{t}$. 
    % Let $x_t \sim \mathcal{N}\bigg(0, \frac{(\beta_{a_t}^2 - 1)\sigma_r^2}{\beta_{a_t}^2}\bigg)$ be a dummy noise variable; this is so that the measurement noise is constant regardless of the action selected for the previous time step. 
    Then by construction:
    \begin{align*}
        y_t = \frac{r_{t} - \mu_{a_{t }}}{\beta_{a_{t}}} = z_{t} + \frac{\epsilon_t(a_t)}{\beta_{a_{t}}} \sim \mathcal{N}(z_{t}, \sigma_r^2)
    \end{align*}
    %%% OLD %%%%
    % \begin{align*}
    %     y_{t} = \frac{r_{t} - \mu_{a_{t }}}{\beta_{a_{t}}} = z_{t} + \frac{\epsilon_t(a)}{\beta_{a_{t}}} \; \sim \mathcal{N}\bigg(z_{t}, \frac{\sigma_r^2}{\beta_{a_{t}}^2}\bigg)
    % \end{align*}
    % which is equivalent to Equation~\ref{lds_measurement_model} as $C\vec{z}_t = z_{t}$.
%     W_t = \begin{bmatrix}
%         0 \\
%         \xi_t \\
%         0 \\
%         \vdots \\
%         0
%     \end{bmatrix} \in \mathbb{R}^{k + 1}
% \end{align*}

\end{proof}

\subsection{Proof of Lemma~\ref{reward_w_tilde_z}}
\label{reward_w_tilde_z_proof}
\begin{proof}
    Recall that due to Assumption~\ref{system_is_observable}, the Kalman gain matrix $K_t$ converges and the steady-state Kalman filter update and prediction can be combined into a single step \citep{gornet2022stochastic}. Equation~\ref{eqn_steady_state_kalman_filter} in our setting becomes:
    \begin{align}
    \label{eqn_tilde_z}
        \Tilde{z}_t = \Gamma \Tilde{z}_{t - 1} + \Gamma K (y_{t - 1} - C \Tilde{z}_{t - 1}) = (\Gamma - \Gamma K C) \Tilde{z}_{t - 1} + \Gamma K y_{t - 1}
    \end{align}
    Since $r_t(a) = \Tilde{r}_t(a) + (r_t(a) - \Tilde{r}_t(a)) = \Tilde{r}_t(a) + \varepsilon_{a; t}$, it suffices to show that $\Tilde{r}_t(a) =$ $\Phi_t(s)^\top \theta_a + b_t(a, s)$.  
    We first show that $\Tilde{r}_t(a) = G_a^\top Y_t + \mu_a + \langle c_a, (\Gamma - \Gamma K C)^s \Tilde{z}_{t - s} \rangle$ where

\begin{align*}
    G_a := \begin{bmatrix}
    c_a^\top(\Gamma - \Gamma K C)^{s - 1} \Gamma K & \cdots & c_a^\top \Gamma K
    \end{bmatrix}^\top \in \mathbb{R}^{s}
    , \;\;\;
    Y_t := \begin{bmatrix}
    y_{t - s} & \cdots & y_{t - 1}
    \end{bmatrix}^\top
    \in \mathbb{R}^{s}
\end{align*}
Recall by definition:
\begin{align*}
    \Tilde{r}_t(a) = \mathbb{E}[r_t(a) | \mathcal{H}_{t - 1}] = \mu_a + c_a^\top\Tilde{z}_t
\end{align*}
    Using Equation~\ref{eqn_tilde_z}, we can continuously unravel $\Tilde{z}_t$ until the $s$th time step before:
    \begin{align*}
         \Tilde{r}_t(a) = c_a^\top(\Gamma - \Gamma K C) \Tilde{z}_{t - 1} + c_a^\top \Gamma K y_{t - 1} + \mu_a
    \end{align*}
    \begin{align*}
        = c_a^\top(\Gamma - \Gamma K C)^2 \Tilde{z}_{t - 2} + c_a^\top(\Gamma - \Gamma K C)\Gamma K y_{t - 2} + c_a^\top \Gamma K y_{t - 1} + \mu_a
    \end{align*}
    \begin{align*}
        = \cdots
    \end{align*}
    \begin{align*}
        = c_a^\top(\Gamma - \Gamma K C)^s \Tilde{z}_{t - s} + c_a^\top(\Gamma - \Gamma K C)^{s - 1} \Gamma K y_{t - s} + \cdots + c_a^\top(\Gamma - \Gamma K C)\Gamma K y_{t - 2} + c_a^\top \Gamma K y_{t - 1} + \mu_a
    \end{align*}
    \begin{align*}
        = G_a^\top Y_t + \mu_a + \langle c_a, (\Gamma - \Gamma K C)^s \Tilde{z}_{t - s} \rangle
    \end{align*}
     \begin{align*}
        = G_a^\top Y_t + \mu_a + b_t(a, s)
    \end{align*}   
    Now let $g_a^j := c_a^\top(\Gamma - \Gamma K C)^j \Gamma K \in \mathbb{R}$. Then:
    \begin{align*}
        G_a^\top Y_t = y_{t - s} g_a^{s - 1} + \cdots + y_{t - 1} g_a^{0}
    \end{align*}
    Using the definition of $y_{t}$,
    \begin{align*}
        G_a^\top Y_t = r_{t - s} \frac{g_a^{s - 1}}{\beta_{a_{t - s}}} + \cdots + r_{t - 1} \frac{g_a^0}{\beta_{a_{t - 1}}} - \frac{\mu_{a_{t - s}} g_a^{s - 1}}{\beta_{a_{t - s}}} - \cdots - \frac{\mu_{a_{t - 1}}g_a^{0}}{\beta_{a_{t - 1}}}
        % + x_{t - s}g_a^{s - 1} + \cdots + x_{t - 1} g_a^0
    \end{align*}
    \begin{align*}
        = R_t^\top \Tilde{\beta}_a - A_t^\top \Tilde{\mu}_a
        % + X_t^\top G_a
    \end{align*}
where
\begin{align*}
    R_t := \begin{bmatrix}
        r_{t - s} e_{a_{t - s}}^\top & \cdots & r_{t - 1} e_{a_{t - 1}}^\top
    \end{bmatrix}^\top \in \mathbb{R}^{s \cdot |\mathcal{A}|}
\end{align*}
\begin{align*}
    \Tilde{\beta}_a = \begin{bmatrix}
        \frac{g_a^{s - 1}}{\beta_{1}} & \cdots & \frac{g_a^{s - 1}}{\beta_{|\mathcal{A}|}} & \cdots & \frac{g_a^{0}}{\beta_{1}} & \cdots & \frac{g_a^{0}}{\beta_{|\mathcal{A}|}}
    \end{bmatrix}^\top \in \mathbb{R}^{s \cdot |\mathcal{A}|}
\end{align*}
\begin{align*}
    A_t := \begin{bmatrix}
        e_{a_{t - s}}^\top & \cdots & e_{a_{t - 1}}^\top
    \end{bmatrix}^\top \in \mathbb{R}^{s \cdot |\mathcal{A}|}, 
\end{align*}
\begin{align*}
        \Tilde{\mu}_a = \begin{bmatrix}
        \frac{\mu_1 g_a^{s - 1}}{\beta_1} & \cdots & \frac{\mu_{|\mathcal{A}|} g_a^{s - 1}}{\beta_{|\mathcal{A}|}} & \cdots & \frac{\mu_1 g_a^{0}}{\beta_{1}} & \cdots & \frac{\mu_{|\mathcal{A}|} g_a^{0}}{\beta_{|\mathcal{A}|}}
    \end{bmatrix}^\top \in \mathbb{R}^{s \cdot |\mathcal{A}|}
\end{align*}
% \begin{align*}
%     X_t := \begin{bmatrix}
%         x_{t - s} & \cdots & x_{t - 1}
%     \end{bmatrix}^\top \in \mathbb{R}^{s}
% \end{align*}

One can verify that by these definitions, for $\theta_a =           \begin{bmatrix}
        \Tilde{\beta}_a  & \Tilde{\mu}_a  & \mu_a
    \end{bmatrix}^\top
    $
\begin{align*}
    \Phi_t(s)^\top \theta_a = \Phi(R_t, A_t)^\top \theta_a = R_t^\top \Tilde{\beta}_a - A_t^\top \Tilde{\mu}_a + \mu_a
\end{align*}
Therefore we have shown:
\begin{align*}
    r_t(a) = \Phi_t(s)^\top \theta_a + b_t(a, s) + \varepsilon_{a; t}
\end{align*}

\end{proof}

The proof for the confidence set lemma (Lemma~\ref{lemma_ellipsoid}) requires lemmas~\ref{bounded_context_lemma} and \ref{noise_process_lemma}. 

% Logic:
% \begin{itemize}
%     \item $r_t$ has Gaussian noise, therefore $r_t$ is sub-Gaussian.
%     \item Because $r_t$ is sub-Gaussian, $r_t \leq C_r$ with probability at least $1 - \delta_r$. We can derive $C_r, \delta_r$.
%     \item $R_t$ has at most $s$ non-zero values and $\|A_t\| = s$
%     \item $\implies$ for a given $t$,  $\|\Phi_t(s)\| \leq L(s) = s \cdot \log (s ) \cdot var(r_t)$
% \end{itemize}
\begin{lemma}
\label{bounded_context_lemma}
    [Bound on Context] Suppose Assumption~\ref{assump_ar_process} holds, then there exists some constant $L(s, \delta_r)$ such that $\|\Phi_t(s)\|^2 \leq L(s, \delta_r)$ for all $t$ with probability at least $1 - \delta_r$, where $\delta_r \in (0, 1)$.
\end{lemma} 
\begin{proof}
    Notice that by construction, $\|R_t\|^2 = R_t^\top R_t = \sum_{j = 1}^s r_{t - j}^2$ and $\|A_t\|^2 = A_t^\top A_t = s$. So:
    \begin{align*}
        \|\Phi_t(s)\|^2 = \sum_{j = 1}^s r_{t - j}^2 + s + 1 
    \end{align*}

    Since every $r_{t - j}$ is Gaussian with variance $\beta_{a_{t - j}}^2\sigma_r^2$, every $r_{t - j}$ is also sub-Gaussian with parameter $\beta_{a_{t - j}}\sigma_r$. Therefore by Lemma~\ref{lemma_sub_gauss_tail_bound}, 
    \begin{align*}
        r_{t - j} < \mu_{a_{t - j}} + \beta_{a_{t - j}} \mathbb{E}[z_{t - j}] + \sqrt{2 \beta_{a_{t - j}}^2\sigma_r^2 \log(1 / \delta_r)})  = R_{\max}(\delta_r)
    \end{align*}
    with probability at least $1 - \delta_r$. Since Assumption~\ref{assump_ar_process} holds, we know that the mean of the AR process $\mathbb{E}[z_{t - j}]$ is always bounded.

    Therefore,
    \begin{align*}
        \|\Phi_t(s)\|^2 < s (R_{\max}(\delta_r)^2 + 1) + 1 = L(s, \delta_r)
    \end{align*}
\end{proof}

For proofs of Lemma~\ref{noise_process_lemma} and Theorem~\ref{thm_regret} we need the following lemma.

\begin{lemma}
    \label{distribution_of_filter_error}
    Let all the information observed up to and including time $t - 1$ be encoded in the filtration $\mathcal{H}_{t - 1} := \sigma(a_1, r_1, ..., a_{t - 1}, r_{t - 1})$ and $\vec{z}_t$ as defined in Equation~\ref{state_evolution} and $\Tilde{z}_t = \mathbb{E}[\vec{z}_t | \mathcal{H}_{t - 1}]$ be the steady-state Kalman filter for $\vec{z}_t$. Then
    $\vec{z}_t - \Tilde{z}_t | \mathcal{H}_{t - 1} \sim \mathcal{N}(\vec{0}, P)$ and 
    $\vec{z}_t - \Tilde{z}_t \sim \mathcal{N}(\vec{0}, P)$
\end{lemma}
\begin{proof}
    First notice that by construction of the steady-state Kalman filter, $\vec{z}_t | \mathcal{H}_{t - 1} \sim \mathcal{N}(\Tilde{z}_t, P)$ and $\Tilde{z}_t | \mathcal{H}_{t - 1}$ is a constant (i.e., not random). Therefore, 
    \begin{align*}
        \vec{z}_t - \Tilde{z}_t | \mathcal{H}_{t - 1} = \vec{z}_t | \mathcal{H}_{t - 1} - \Tilde{z}_t | \mathcal{H}_{t - 1} \sim \mathcal{N}(\vec{0}, P)
    \end{align*}
     Since $\mathcal{N}(\vec{0}, P)$ is a fixed distribution and $P$ does not depend on $\mathcal{H}_{t - 1}$, this implies that $\vec{z}_t - \Tilde{z}_t \sim \mathcal{N}(\vec{0}, P)$ as well.
\end{proof}

\begin{lemma}
\label{noise_process_lemma}
    [Noise Process Property] 
    % Let the filtration be $\mathcal{F}_{t} := \{\mathcal{F}_0, \Phi(s)_1, a_1, r_1, ..., \Phi(s)_{t - 1}, a_{t - 1},r_{t - 1}, \Phi_t(s)\}$ where $\mathcal{F}_0$ contains any prior knowledge. 
    Let all the information observed up to and including time $t - 1$ be encoded in the filtration $\mathcal{H}_{t - 1} := \sigma(a_1, r_1, ..., a_{t - 1}, r_{t - 1})$.
    For any given $a$, the noise process
    $\{\varepsilon_{a; t}\}_t$ from Equation~\ref{eqn_reward_wrt_tilde_z} is a martingale difference sequence given filtration $\mathcal{H}_{t - 1}$ and is conditionally $R$-subgaussian for some constant $R \geq 0$,
    \begin{gather*}
        \forall t \geq 1, \mathbb{E}[\varepsilon_{a; t} | \mathcal{H}_{t - 1}] = 0 \\
        \forall \alpha \in \mathbb{R}, \mathbb{E}[e^{\alpha \varepsilon_{a; t}} | \mathcal{H}_{t - 1}] \leq \text{exp}(\alpha^2 R^2 / 2)
    \end{gather*}
\end{lemma}
% $$
%     \varepsilon_{a; t} := r_t(a) - \Tilde{r}_t(a) = \langle c_a, \vec{z}_t - \Tilde{z}_t \rangle + \epsilon_t(a) \\
%     \sim \mathcal{N}(0, c_a^\top P c_a + \sigma_r^2)
% $$
\begin{proof}
Fix some $a \in \mathcal{A}$. We first show that $\mathbb{E}[\varepsilon_{a; t} | \mathcal{H}_{t - 1}] = 0$.
\begin{align*}
    \mathbb{E}[\varepsilon_{a; t} | \mathcal{H}_{t - 1}] =  \mathbb{E}[\langle c_a, \vec{z}_t - \Tilde{z}_t \rangle + \epsilon_t(a) | \mathcal{H}_{t - 1}]
\end{align*}
\begin{align*}
    = c_a^\top (\mathbb{E}[\vec{z}_t - \Tilde{z}_t | \mathcal{H}_{t - 1}]) + \mathbb{E}[\epsilon_t(a) | \mathcal{H}_{t - 1}]
\end{align*}
\begin{align*}
    = c_a^\top \mathbb{E}[\vec{z}_t - \Tilde{z}_t | \mathcal{H}_{t - 1}] \;\;\; \text{($\epsilon_t(a)$ is independent of $\mathcal{H}_{t - 1}$ and $\mathbb{E}[\epsilon_t(a)] = 0$)}
\end{align*}
\begin{align*}
    = c_a^\top \vec{0} = 0 \;\;\; \text{(Lemma~\ref{distribution_of_filter_error})}
\end{align*}

% Next, we show that $\varepsilon_{a; t} \sim \mathcal{N}(0, c_a^\top P c_a + \sigma_r^2)$.
% \begin{align*}
%     \mathbb{E}[\varepsilon_{a; t}] = \mathbb{E}[\langle c_a, \vec{z}_t - \Tilde{z}_t \rangle + \epsilon_t(a)]
%     = c_a^\top(\mathbb{E}[\vec{z}_t - \Tilde{z}_t]) + \mathbb{E}[\epsilon_t(a)]
% \end{align*}
% \begin{align*}
%     = c_a^\top \vec{0} = 0 \;\;\; \text{(Lemma~\ref{distribution_of_filter_error})}
% \end{align*}

% \begin{align*}
%     \text{Var}[\varepsilon_{a; t}] = \text{Var}[\langle c_a, \vec{z}_t - \Tilde{z}_t \rangle + \epsilon_t(a)]
% \end{align*}
% \begin{align*}
%     = \text{Var}[\langle c_a, \vec{z}_t - \Tilde{z}_t \rangle] + \text{Var}[\epsilon_t(a)] \;\;\; \text{($\epsilon_t(a)$ is independent of $\vec{z}_t, \Tilde{z}_t$)}
% \end{align*}
% \begin{align*}
%     = c_a^\top \text{Cov}(\vec{z}_t - \Tilde{z}_t) c_a + \sigma_r^2
% \end{align*}
% \begin{align*}
%     = c_a^\top P c_a + \sigma_r^2 \;\;\; \text{(Lemma~\ref{distribution_of_filter_error})}
% \end{align*}

To prove that $\varepsilon_{a; t}$ is conditionally $R$-subgaussian for some $R$, we first show that $\varepsilon_{a; t} | \mathcal{H}_{t - 1} \sim \mathcal{N}(0, c_a^\top P c_a + \sigma_r^2)$. Notice that $c_a^\top (\vec{z}_t - \Tilde{z}_t) | \mathcal{H}_{t - 1} \sim \mathcal{N}(0, c_a^\top P c_a)$ by Lemma~\ref{distribution_of_filter_error} and $\epsilon_t(a) | \mathcal{H}_{t - 1} \sim \mathcal{N}(0, \sigma_r^2)$ since $\epsilon_t(a)$ is independent of $\mathcal{H}_{t - 1}$. Therefore:
\begin{align*}
    \varepsilon_{a; t} | \mathcal{H}_{t - 1} = c_a^\top (\vec{z}_t - \Tilde{z}_t) | \mathcal{H}_{t - 1} + \epsilon_t(a) | \mathcal{H}_{t - 1} \sim \mathcal{N}(0, c_a^\top P c_a + \beta_{a}^2\sigma_r^2)
\end{align*}
The moment generating function (MGF) for the normal random variable $\varepsilon_{a; t}$ is:
\begin{align*}
    M_{\varepsilon_{a; t}}(\alpha) = \mathbb{E}[e^{\alpha \varepsilon_{a; t}}] = \text{exp}(\alpha^2 (c_a^\top P c_a + \beta_{a}^2\sigma_r^2)  /2) \;\; \forall \alpha \in \mathbb{R}
\end{align*}
Then:
\begin{align*}
    \mathbb{E}[e^{\alpha \varepsilon_{a; t}} | \mathcal{H}_{t - 1}] = \mathbb{E}[e^{\alpha \varepsilon_{a; t}}] \;\;\; \text{(as shown above, $\varepsilon_{a; t}$ is independent of $\mathcal{H}_{t - 1}$)}
\end{align*}
\begin{align*}
    = \text{exp}(\alpha^2(c_a^\top P c_a + \beta_{a}^2\sigma_r^2) /2)
\end{align*}
\begin{align*}
    \leq \text{exp}(\alpha^2 R^2 / 2) \;\; \forall \alpha \in \mathbb{R}
\end{align*}
for some $R^2 \geq c_a^\top P c_a + \beta_{a}^2\sigma_r^2$
\end{proof}

\subsection{Proof of Lemma~\ref{lemma_ellipsoid}}
\label{confidence_ellipsoid_proof}
\begin{proof}
First notice that for all actions $a$,
\begin{align*}
    \hat{\theta}_{a, t} - \theta_a = V_{a, t}^{-1} \sum_{j = 1}^t \mathbb{I}[a_j = a] \Phi_j(s) r_j - \theta_a
\end{align*}
\begin{align*}
    = V_{a, t}^{-1} \sum_{j = 1}^t \mathbb{I}[a_j = a] \Phi_j(s) (\Phi_j(s)^\top \theta_a + b_j(a, s) + \varepsilon_{a; t}) - \theta_a
\end{align*}
\begin{align*}
    = V_{a, t}^{-1} \sum_{j = 1}^t \mathbb{I}[a_j = a] \Phi_j(s) (\Phi_j(s)^\top \theta_a + b_j(a, s) + \varepsilon_{a; t}) - V_{a, t}^{-1} (\lambda I + \sum_{j = 1}^t \mathbb{I}[a_j = a] \Phi_j(s) \Phi_j(s)^\top) \theta_a
\end{align*}
\begin{align*}
    = V_{a, t}^{-1} \sum_{j = 1}^t \mathbb{I}[a_j = a] \Phi_j(s) b_j(a, s) + V_{a, t}^{-1} \sum_{j = 1}^t \mathbb{I}[a_j = a] \Phi_j(s) \varepsilon_{a; j} - \lambda V_{a, t}^{-1} \theta_a
\end{align*}
For the first term,
\begin{align*}
    \bigg\|V_{a, t}^{-1} \sum_{j = 1}^t \mathbb{I}[a_j = a] \Phi_j(s) b_j(a, s) \bigg\|_{V_{a, t}} = \bigg\| \sum_{j = 1}^t \mathbb{I}[a_j = a] \Phi_j(s) b_j(a, s) \bigg\|_{V_{a, t}^{-1}}
\end{align*}
\begin{align*}
    \leq \sum_{j = 1}^t \mathbb{I}[a_j = a] \| \Phi_j(s) b_j(a, s) \|_{V_{a, t}^{-1}} \;\;\; \text{(generalized triangle-inequality)}
\end{align*}
\begin{align*}
    = \sum_{j = 1}^t \mathbb{I}[a_j = a] \sqrt{b_j(a, s)^2 \Phi_j(s) ^\top V_{a, t}^{-1} \Phi_j(s)}
\end{align*}
\begin{align*}
    \leq \sqrt{\sum_{j = 1}^t \mathbb{I}[a_j = a] b_j(a, s)^2} \sqrt{\sum_{j = 1}^t \mathbb{I}[a_j = a] \Phi_j(s)^\top V_{a, t}^{-1} \Phi_j(s)} \;\;\; \text{(Cauchy-Schwartz)}
\end{align*}
\begin{align*}
    = \tau(a, s)_t \sqrt{\sum_{j = 1}^t \mathbb{I}[a_j = a] b_j(a, s)^2}
\end{align*}
where $\tau(a, s)_t = \sqrt{\sum_{j = 1}^t \mathbb{I}[a_j = a] \|\Phi_j(s)\|_{V_{a, t}^{-1}}^2}$

For the second term, 
\begin{align*}
    \bigg \|V_{a, t}^{-1} \sum_{j = 1}^t \mathbb{I}[a_j = a] \Phi_j(s) \varepsilon_{a; j} \bigg \|_{V_{a, t}} = \bigg \|\sum_{j = 1}^t \mathbb{I}[a_j = a] \Phi_j(s) \varepsilon_{a; j} \bigg \|_{V_{a, t}^{-1}}
\end{align*}

By Lemma~\ref{noise_process_lemma}, since the noise process $\varepsilon_{a; j}$ satisfies the assumptions of Theorem~\ref{normalized_bound_for_martingales},
\begin{align*}
    \bigg \|\sum_{j = 1}^t \mathbb{I}[a_j = a] \Phi_j(s) \varepsilon_{a; j} \bigg \|_{V_{a, t}^{-1}} \leq \sqrt{2 R^2 \log \bigg (\frac{\text{det}(V_{a, t})^{1/2} \text{det}(\lambda I)^{-1/2}}{\delta_\beta}\bigg)}
\end{align*}
with probability at-least $1 - \delta_\beta$.
Using Lemma~\ref{determinant_trace_ineq} (determinant-trace inequality),
\begin{align*}
    \text{det}(V_{a, t}) \leq \bigg( \lambda + \frac{n_{a, t} L(s, \delta_r)}{2s|\mathcal{A}| + 1}\bigg)^{2s|\mathcal{A}| + 1}
\end{align*}
where $n_{a, t} := \sum_{j = 1}^t \mathbb{I}[a_j = a]$ and $\|\Phi_j(s)\|^2 \leq L(s, \delta_r)$ for all $j \in [t]$ because of Lemma~\ref{bounded_context_lemma}. 
\begin{align*}
    \implies \bigg \|\sum_{j = 1}^t \mathbb{I}[a_j = a] \Phi_j(s) \varepsilon_{a; j} \bigg \|_{V_{a, t}^{-1}} \leq R \sqrt{(2s|\mathcal{A}| + 1) \log \bigg (\frac{1 + n_{a, t} L(s, \delta_r) / \lambda}{\delta_\beta}\bigg)}
\end{align*}

For the third term,

\begin{align*}
    \|\lambda V_{a, t}^{-1} \theta_a\|_{V_{a, t}} = \lambda \| \theta_a \|_{V_{a, t}^{-1}} \leq \sqrt{\lambda} \| \theta_a \| \leq \sqrt{\lambda} S_a
\end{align*}
since $\| \theta_a \|^2_{V_{a, t}^{-1}} \leq \frac{1}{\lambda_{\min}(V_{a, t})} \| \theta_a \|^2 \leq \frac{1}{\lambda} \| \theta_a \|^2$ and using Assumption~\ref{reward_param_assumption}.

Using generalized triangle inequality
\begin{multline*}
     \implies \|\hat{\theta}_{a, t} - \theta_a \|_{V_{a, t}} \leq \bigg \|V_{a, t}^{-1} \sum_{j = 1}^t \mathbb{I}[a_j = a] \Phi_j(s) b_j(a, s) \bigg \|_{V_{a, t}} \\
     + \bigg \| V_{a, t}^{-1} \sum_{j = 1}^t \mathbb{I}[a_j = a] \Phi_j(s) \varepsilon_{a; j} \bigg \|_{V_{a, t}} + \bigg \| \lambda V_{a, t}^{-1} \theta_a \bigg \|_{V_{a, t}}
\end{multline*}
\begin{align*}
    \leq R \sqrt{(2s|\mathcal{A}| + 1) \log \bigg (\frac{1 + n_{a, t} L(s, \delta_r) / \lambda}{\delta_\beta}\bigg)} + \sqrt{\lambda} S_a + \tau(a, s)_t \sqrt{\sum_{j = 1}^t \mathbb{I}[a_j = a] b_j(a, s)^2}
\end{align*}
Finally, for readability, we let $\delta_{\beta} = \delta_{r} = \delta / 2$. Therefore with probability at least $1 - \delta$, 
\begin{multline*}
    \|\hat{\theta}_{a, t} - \theta_a \|_{V_{a, t}} \leq \\
    R \sqrt{(2s|\mathcal{A}| + 1) \log \bigg (\frac{1 + n_{a, t} L(s, \delta/2) / \lambda}{\delta/2}\bigg)} + \sqrt{\lambda} S_a + \tau(a, s)_t \sqrt{\sum_{j = 1}^t \mathbb{I}[a_j = a] b_j(a, s)^2} 
\end{multline*}
\end{proof}

\subsection{Proof of Theorem~\ref{thm_regret}}
\label{regret_proof}
\begin{proof}
    Using Assumption~\ref{reward_param_assumption} with Lemma~\ref{lemma_ellipsoid}, it suffices to prove the bound on the event that true parameter $\theta_a \in \mathcal{C}_{a, t}$ (Equation~\ref{our_confidence_set}) for $\forall a \in \mathcal{A}$ and $t \in [T]$.  
Recall the regret in our setting (Equation~\ref{eqn_regret}) is:
\begin{align*}
    \text{Regret}(T; \pi) =
    \sum_{t = 1}^T  \mathbb{E}[r_t(a_t^*) - r_t(a_t) | \vec{z}_t]
\end{align*}
where $a^*_t$ is the optimal action and $a_t$ is the action selected by Algorithm~\ref{alg_latent_ar_ucb} at time step $t$.

To assist with the proof, we consider an intermediate agent that knows the ground-truth parameters and therefore has the exact steady-state Kalman filter prediction $\Tilde{z}_t$ and selects actions $\Tilde{a}_t = \underset{a}{\arg \max} \;\; \Tilde{r}_t(a)$.

Let $\Delta_t := \mathbb{E}[r_t(a_t^*) - r_t(a_t) | \vec{z}_t]$ denote the instantaneous regret at time $t$. Then:
\begin{align*}
    \Delta_t = c_{a^*_t}^\top \vec{z}_t + \mu_{a^*_t} -c_{a_t}^\top \vec{z}_t - \mu_{a_t}   
\end{align*}
\begin{align*}
    = (c_{a^*_t}^\top \vec{z}_t + \mu_{a^*_t} - c_{\Tilde{a}_t}^\top \Tilde{z}_t - \mu_{\Tilde{a}_t}) + (c_{\Tilde{a}_t}^\top \Tilde{z}_t + \mu_{\Tilde{a}_t} - c_{a_t}^\top \vec{z}_t - \mu_{a_t})   
\end{align*}
First notice that:
\begin{align*}
    c_{a^*_t}^\top \vec{z}_t + \mu_{a^*_t} - c_{\Tilde{a}_t}^\top \Tilde{z}_t - \mu_{\Tilde{a}_t} = c_{a^*_t}^\top \vec{z}_t - c_{a^*_t}^\top \Tilde{z}_t + c_{a^*_t}^\top \Tilde{z}_t + \mu_{a^*_t} - c_{\Tilde{a}_t}^\top \Tilde{z}_t - \mu_{\Tilde{a}_t}
\end{align*}
\begin{align*}
    = c_{a^*_t}^\top (\vec{z}_t - \Tilde{z}_t) + \Tilde{r}_t(a^*_t) - \Tilde{r}_t(\Tilde{a}_t)
\end{align*}
\begin{align*}
    \leq c_{a^*_t}^\top (\vec{z}_t - \Tilde{z}_t)
\end{align*}
since $\Tilde{r}_t(a^*_t) - \Tilde{r}_t(\Tilde{a}_t) \leq 0$ by the action-selection strategy of the intermediate agent.

Next notice that:
\begin{align*}
    c_{\Tilde{a}_t}^\top \Tilde{z}_t + \mu_{\Tilde{a}_t} - c_{a_t}^\top \vec{z}_t - \mu_{a_t} = c_{\Tilde{a}_t}^\top \Tilde{z}_t + \mu_{\Tilde{a}_t} + (c_{a_t}^\top \Tilde{z}_t - c_{a_t}^\top \Tilde{z}_t) - c_{a_t}^\top \vec{z}_t - \mu_{a_t}
\end{align*}
\begin{align*}
    = \Tilde{r}_t(\Tilde{a}_t) - \Tilde{r}_t(a_t) - c_{a_t}^\top (\vec{z}_t - \Tilde{z}_t)
\end{align*}
\begin{align*}
    \implies \Delta_t \leq  (c_{a^*_t} - c_{a_t})^\top (\vec{z}_t - \Tilde{z}_t) +  \Tilde{r}_t(\Tilde{a}_t) - \Tilde{r}_t(a_t)
\end{align*}
We first focus on $\Tilde{r}_t(\Tilde{a}_t) - \Tilde{r}_t(a_t)$. Recall by Lemma~\ref{reward_w_tilde_z} that $\Tilde{r}_t(a) = \Phi_t(s)^\top \theta_a + b_t(a, s)$. Then:
\begin{align*}
    \Tilde{r}_t(\Tilde{a}_t) - \Tilde{r}_t(a_t) =  \Phi_t(s)^\top \theta_{\Tilde{a}_t} + b_t(\Tilde{a}_t, s) -  \Phi_t(s)^\top \theta_{a_t} - b_t(a_t, s)
\end{align*}
Now let $\theta'_{a, t} = \max \{\mathcal{C}_{a, t - 1}\}$ denote the max value of the confidence set $\mathcal{C}_{a, t - 1}$ constructed at time $t$. Notice that:
\begin{align*}
    \Phi_t(s)^\top \theta_{\Tilde{a}_t} \leq \Phi_t(s)^\top \theta'_{\Tilde{a}_t, t} \leq \Phi_t(s)^\top \theta'_{a_t, t}
\end{align*}
where the first inequality is because $\theta_{\Tilde{a}_t} \in \mathcal{C}_{\Tilde{a}_t, t - 1}$ and $\theta'_{\Tilde{a}_t, t} = \max \{\mathcal{C}_{\Tilde{a}_t, t - 1}\}$, and the second inequality is by the action-selection strategy of Algorithm~\ref{alg_latent_ar_ucb} (i.e.,  $a_t = \underset{a}{\arg \max} \;\; \Phi_t(s)^\top \theta'_{a, t}$).
$\implies$
\begin{align*}
     \Tilde{r}_t(\Tilde{a}_t) - \Tilde{r}_t(a_t) \leq \Phi_t(s)^\top(\theta'_{a_t, t} - \theta_{a_t}) + b_t(\Tilde{a}_t, s) - b_t(a_t, s)
\end{align*}
\begin{multline*}
    \leq \|\Phi_t(s)\|_{V_{a, t - 1}^{-1}} \| \theta'_{a_t, t} - \theta_{a_t}\|_{V_{a_t, t - 1}} + b_t(\Tilde{a}_t, s) - b_t(a_t, s) \\
    \text{(by Cauchy-Schwartz and $\|\cdot\|_{V_{a_t, t - 1}^{-1}} \leq \| \cdot\|_{V_{a, t - 1}}$)}
\end{multline*}
% the max. distance of two points in the confidence set is 2 times the radius
\begin{align*}
    \leq 2  \|\Phi_t(s)\|_{V_{a_t, t - 1}^{-1}} \beta_{a_t, t - 1}(\delta') + b_t(\Tilde{a}_t, s) - b_t(a_t, s)
\end{align*}
\begin{align*}
    \leq 2  \|\Phi_t(s)\|_{V_{a_t, t - 1}^{-1}} \beta_{a_t, t - 1}(\delta') + 2 \max_a |b_t(a, s)|
\end{align*}

We now focus on $(c_{a^*_t} - c_{a_t})^\top (\vec{z}_t - \Tilde{z}_t)$. 

First,
\begin{align*}
    (c_{a^*_t} - c_{a_t})^\top (\vec{z}_t - \Tilde{z}_t) \leq \|c_{a^*_t} - c_{a_t}\| \|\vec{z}_t - \Tilde{z}_t\| \;\;\; \text{(Cauchy-Schwartz)}
\end{align*}
\begin{align*}
    \leq 2 \max_a \|c_a\| \|\vec{z}_t - \Tilde{z}_t\|
\end{align*}

By Lemma~\ref{distribution_of_filter_error}, we know that $\vec{z}_t - \Tilde{z}_t \sim \mathcal{N}_{k}(\vec{0}, P)$. By Lemma~\ref{normal_is_sub_gaussian_lemma}, $\vec{z}_t - \Tilde{z}_t$ is sub-Gaussian with parameter $\sigma^2 = \| P\|_{\text{op}}$. Finally by Theorem~\ref{norm_of_sub_gauss_rv}, with probability at least $1 - \delta_z$ for $\delta_z \in (0, 1)$,
\begin{align*}
    \| \vec{z}_t - \Tilde{z}_t \| \leq 4 \sqrt{\|P\|_{\text{op}} k} + 2 \sqrt{\|P\|_{\text{op}} \log (1 / \delta_z)}
\end{align*}
\begin{align*}
    \leq 4 \sqrt{\|P\|_{\text{op}}} (\sqrt{k} + \sqrt{\log (1 / \delta_z)})
\end{align*}
\begin{align*}
    \leq 4 \sqrt{\|P\|_{\text{op}}} \sqrt{2 (k + \log (1 / \delta_z))} \;\;\; \text{($\sqrt{a} + \sqrt{b} \leq \sqrt{2(a + b)}$ for $a, b \in \mathbb{R}_{\geq 0}$)}
\end{align*}

We bound $\|P\|_{\text{op}}$. Recall that for the steady-state Kalman filter, $P = \Gamma P \Gamma^\top + W - \Gamma P C^\top (CPC^\top + V)^{-1}CP\Gamma^\top$.

Using triangle inequality,
\begin{align*}
    \|P\|_{\text{op}} = \|\Gamma P \Gamma^\top - \Gamma P C^\top (CPC^\top + V)^{-1}CP\Gamma^\top \|_{\text{op}} + \|W\|_{\text{op}}
\end{align*}

We now show that $\|\Gamma P \Gamma^\top - \Gamma P C^\top (CPC^\top + V)^{-1}CP\Gamma^\top \|_{\text{op}} \leq \|\Gamma P \Gamma^\top\|_{\text{op}}$. To do so, we show that the following three matrices are positive semi-definite (PSD): (1) $A = \Gamma P \Gamma^\top$, (2) $B = \Gamma P C^\top (CPC^\top + V)^{-1}CP\Gamma^\top$, and (3) $A - B$.

$A$ is PSD because $P$ is PSD and using Lemma~\ref{ABA_is_PSD_lemma}.

We now show $B$ is PSD. Consider some vector $v \in \mathbb{R}^{k}$. 
Then $v^\top B v = v^\top (\Gamma P C^\top (CPC^\top + V)^{-1}CP\Gamma^\top) v = (CP\Gamma^\top v)^\top (CPC^\top + V)^{-1}CP\Gamma^\top v = (CP\Gamma^\top v)^2 (CPC^\top + V)^{-1}$.

Now since $C = \begin{bmatrix}
    1 & 0 & \cdots & 0
\end{bmatrix} \in \mathbb{R}^{1 \times k}$, $C P C^\top = P_{11} \geq 0$ since $P$ is PSD so diagonal entries are non-negative. Also $V = \frac{\sigma_r^2}{\min_a \beta_a^2} \implies (CPC^\top + V)^{-1} = \frac{1}{P_{11} + \frac{\sigma_r^2}{\min_a \beta_a^2}} \geq 0$.

We now show $A - B$ is PSD. Since $P$ is PSD we know that there exists a PSD matrix $P^{1/2}$ such that $P = P^{1/2}P^{1/2}$. Therefore $A - B = \Gamma P \Gamma^\top - \Gamma P C^\top (CPC^\top + V)^{-1}CP\Gamma^\top = \Gamma P^{1/2} (I - P^{1/2}C^\top(C P^{1/2} P^{1/2} C^\top + V)^{-1} CP^{1/2}) P^{1/2} \Gamma^\top$ = $\Gamma P^{1/2} (I - \nu(\nu^\top \nu + V)^{-1} \nu^\top) P^{1/2} \Gamma^\top$ for $\nu = P^{1/2}C^\top \in \mathbb{R}^{k}$.

Notice that $\nu(\nu^\top \nu + V)^{-1} \nu^\top = \frac{\nu \nu^\top}{\nu^\top \nu + V}$
$= \frac{\frac{\nu}{\|\nu\|} \frac{\nu^\top}{\|\nu\|}}{\frac{v^\top v}{\|\nu\|^2} + \frac{V}{\|\nu\|^2}}$
$= \frac{\nu'\nu'^\top}{1 + \frac{V}{\|\nu\|^2}}$ where $\nu' = \nu / \|\nu\|$. Since $\nu'$ is a unit vector, $\nu' \nu'^\top$ has only one non-zero eigenvalue which is 1, which implies $\frac{\nu'\nu'^\top}{1 + \frac{V}{\|\nu\|^2}}$ has one non-zero eigenvalue which is $\frac{1}{1 + \frac{V}{\|\nu\|^2}}$. Furthermore this implies that $I - \frac{\nu'\nu'^\top}{1 + \frac{V}{\|\nu\|^2}}$ has eigenvalues either $1$ or $1 - \frac{1}{1 + \frac{V}{\|\nu\|^2}} \geq 0$. Therefore, $I - \nu(\nu^\top \nu + V)^{-1} \nu^\top$ is PSD and so is $\Gamma P^{1/2} (I - \nu(\nu^\top \nu + V)^{-1} \nu^\top) P^{1/2} \Gamma^\top$ by Lemma~\ref{ABA_is_PSD_lemma}.

Since we have shown $A$, $B$, and $A - B$ are PSD, then by Lemma~\ref{op_norm_diff_inequality}, $\|\Gamma P \Gamma^\top - \Gamma P C^\top (CPC^\top + V)^{-1}CP\Gamma^\top \|_{\text{op}} \leq \|\Gamma P \Gamma^\top\|_{\text{op}}$.

$\implies$
\begin{align*}
        \|P\|_{\text{op}} \leq \|\Gamma P \Gamma^\top \|_{\text{op}} + \|W\|_{\text{op}}
\end{align*}
Now, 
% ref: https://math.stackexchange.com/questions/2415270/how-to-give-an-upper-bound-of-the-maximum-eigenvalue-of-a-b-at
\begin{align*}
    \|\Gamma P \Gamma^\top \|_{\text{op}} \leq \|\Gamma\|_{\text{op}} \|P\|_{\text{op}} \|\Gamma^\top\|_{\text{op}} = \sigma_{\max}(\Gamma)^2  \|P\|_{\text{op}}
\end{align*}
since $A$ and $A^\top$ have the same singular values for any matrix $A$.

Also in our setting, $W$ is a matrix of all 0s except for $\sigma_z^2$ in the first diagonal entry, so $\|W\|_{\text{op}} = \sigma_z^2$. So,
\begin{align*}
    \|P\|_{\text{op}} \leq \sigma_{\max}(\Gamma)^2  \|P\|_{\text{op}} + \sigma_z^2 \implies \|P\|_{\text{op}} \leq \frac{\sigma_z^2}{1 - \sigma_{\max}(\Gamma)^2}
\end{align*}

Therefore, at every $t$, the instantaneous regret is bounded as so:
\begin{multline*}
    \Delta_t \leq 8 \max_a \|c_a\| \sqrt{\frac{\sigma_z^2}{1 - \sigma_{\max}(\Gamma)^2}} \sqrt{2(k + \log(1 / \delta_z))} \\ + 2 \|\Phi_t(s)\|_{V_{a_t, t - 1}^{-1}} \beta_{a_t, t - 1}(\delta') + 2 \max_a |b_t(a, s)|
\end{multline*}
So,
\begin{align*}
    \text{Regret}(T; \pi_{\text{LARL}}) = \sum_{t = 1}^T \Delta_t 
\end{align*}
\begin{multline*}
    \leq 8 \max_a \|c_a\| \sqrt{\frac{\sigma_z^2}{1 - \sigma_{\max}(\Gamma)^2}} \sqrt{2(k + \log(1 / \delta_z))} \cdot T \\
    + 2 \sum_{t = 1}^T \|\Phi_t(s)\|_{V_{a_t, t - 1}^{-1}} \beta_{a_t, t - 1}(\delta') + 2 \sum_{t = 1}^T \max_a |b_t(a, s)|
\end{multline*}
Now,
\begin{align*}
    2 \sum_{t = 1}^T \|\Phi_t(s)\|_{V_{a_t, t - 1}^{-1}} \beta_{a_t, t - 1}(\delta') \leq 2 \beta_{T}(\delta') \sqrt{\sum_{t = 1}^T \|\Phi_t(s)\|_{V_{a_t, t - 1}^{-1}}^2} \sqrt{T} 
\end{align*}
where $\beta_{T}(\delta') = \max_a \beta_{a, T - 1}(\delta')$ and $\sum_{t = 1}^T \|\Phi_t(s)\|_{V_{a_t, t - 1}^{-1}} \leq \sqrt{T \cdot \sum_{t = 1}^T \|\Phi_t(s)\|_{V_{a_t, t - 1}^{-1}}^2}$ by variant of Cauchy-Schwartz.

For readability, let $\delta_z = \delta_b = \delta_r = \delta / 3$. Then w.p. at-least $1 - \delta$,
\begin{gather*}
    \text{Regret}(T; \pi_{\text{LARL}}) \leq 8 \max_a \|c_a\| \sqrt{\frac{\sigma_z^2}{1 - \sigma_{\max}(\Gamma)^2}} \sqrt{2(k + \log(3 / \delta))} \cdot T \\
    + 2 \beta_{T}(2\delta/3) \sqrt{\sum_{t = 1}^T \|\Phi_t(s)\|_{V_{a_t, t - 1}^{-1}}^2} \sqrt{T} + 2 \sum_{t = 1}^T \max_a |b_t(a, s)|
\end{gather*}

\end{proof}

\section{Auxillary Theorems and Lemmas}
% \begin{theorem}
% \label{thm_confidence_ellipsoid}
% [Confidence Ellipsoid] (Thm. 2 in \cite{abbasi2011improved})
%     For any $\delta \in (0, 1),$ under Assumptions~\ref{action_set_assumption}, \ref{reward_param_assumption}, \ref{standard_noise_assumption},
%     for any $\mathcal{F}_t$-adapted sequence $x_1,...,x_t$, the regularized least squares estimator $\theta_t^{\text{RLS}} = (\lambda I + \sum_{j = 1}^t x_j x_j^\top )^{-1} \sum_{j = 1}^t r_j x_j$ is such that for any fixed $t \geq 1$:
%     \begin{equation}
%         \|\theta_t^{\text{RLS}} - \theta^*\|_{V_t} \leq \beta_t(\delta)
%     \end{equation}
%     % \begin{equation*}
%     %     \forall x \in \mathbb{R}^d, \; |x^{\top}(\hat{\theta}_t - \theta^*)| \leq \|x\|_{V_t^{-1}} \beta_t(\delta)
%     % \end{equation*}
%     w.p. $1 - \delta$ where the randomness is w.r.t. the noise and any source of randomization in the policy,
%     where
%     %%% OLD ONE %%%
%     % \begin{equation}
%     % \label{eqn_radius}
%     %     \beta_t(\delta) = R \sqrt{2 \log \frac{(\lambda + t)^{d/ 2} \lambda^{-d/2}}{\delta}} + \sqrt{\lambda} S
%     % \end{equation}
%     %%% VERBATIM FROM AY-2011 PAPER %%%
%     \begin{equation}
%     \label{eqn_radius}
%         \beta_t(\delta) = R \sqrt{d \log \bigg(\frac{1 + tA^2 / \lambda}{\delta}}\bigg) + \sqrt{\lambda} S
%     \end{equation}
% \end{theorem}

\begin{theorem}
    (Theorem 1 in \cite{abbasi2011improved})
\label{normalized_bound_for_martingales}
Let $\{\mathcal{F}_t\}_{t = 0}^\infty$ be a filtration. Let $\{\eta_t\}_{t = 1}^\infty$ be a real-valued stochastic process such that $\eta_t$ is $\mathcal{F}_t$-measurable and $\eta_t$ is conditionally R-sub-Gaussian for some $R \geq 0$. Namely:
\begin{equation*}
    \forall \; \lambda \in \mathbb{R} \;\;\; \mathbb{E}[e^{\lambda \eta_t} | \mathcal{F}_{t - 1}] \leq \exp \bigg( \frac{\lambda^2 R^2}{2}\bigg)
\end{equation*}

Let $\{x_t\}_{t = 1}^\infty$ be an $\mathbb{R}^d$-valued stochastic process such that $x_t$ is $\mathcal{F}_{t - 1}$-measurable. Assume that $V_0 \in \mathbb{R}^{d \times d}$ is a positive definite matrix. For any $t \geq 1$, define:
\begin{equation*}
    V_t = V_0 + \sum_{j = 1}^t x_j x_j^\top \;\;\; S_t = \sum_{j = 1}^t \eta_j x_j
\end{equation*}

Then for any $\delta > 0$, with probability atleast $1 - \delta$, for all $t \geq 1$,
\begin{equation*}
    \|S_t\|^2_{V_t^{-1}} \leq 2 R^2 \log \bigg(\frac{\text{det}(V_t)^{1/2} \text{det}(V_0)^{-1/2}}{\delta}\bigg)
\end{equation*}
\end{theorem}

\begin{lemma}
    (Lemma 10 Determinant-Trace Inequality in \cite{abbasi2011improved})
    \label{determinant_trace_ineq}
    Suppose $x_1,...,x_t \in \mathbb{R}^d$ and $\|x_j\| \leq L \; \forall \; j \in [t]$. Let $V_{t} = \lambda I + \sum_{j = 1}^t x_j x_j^\top$ for some $\lambda > 0$. Then:
    \begin{equation*}
        \text{det}(V_t) \leq \bigg(\lambda + \frac{tL^2}{d} \bigg)^d
    \end{equation*}
\end{lemma}
\begin{lemma}
\label{normal_is_sub_gaussian_lemma}
    (Lemma 8.2 in \cite{subgaussianlecnotes2019}) Let $X \in \mathbb{R}^d$ be a random vector that is normally distributed $X \sim \mathcal{N}(0, \Sigma)$. Then $X$ is a sub-Gaussian random vector with parameter $\| \Sigma\|_{\text{op}}$.
\end{lemma}

\begin{theorem}
\label{norm_of_sub_gauss_rv}
    (Theorem 8.3 in \cite{subgaussianlecnotes2019}) $X \in \mathbb{R}^d$ be a sub-Gaussian random vector with parameter $\sigma^2$, then with probability at least $1 - \delta$ for $\delta \in (0, 1)$:
    \begin{align*}
        \|X\| \leq 4 \sigma \sqrt{d} + 2 \sigma \sqrt{\log (1 / \delta)}
    \end{align*}
\end{theorem}

% Prop. 6.6 from https://www.stats.ox.ac.uk/~rebeschi/teaching/AFoL/22/material/lecture06.pdf
\begin{lemma}
    \label{lemma_sub_gauss_tail_bound}
    (Sub-Gaussian upper tail bound) Let $X$ be sub-Gaussian with variance proxy $\sigma^2$. Then for any $\delta \in [0, 1]$ we have:
    \begin{align*}
        \mathbb{P}(X - \mathbb{E}[X] < \sqrt{2 \sigma^2 \log(1 / \delta)}) \leq 1 - \delta
    \end{align*}
\end{lemma}

%%% HELPFUL PSD LEMMAS %%%
\begin{lemma}
\label{ABA_is_PSD_lemma}
    If $B \in \mathbb{R}^{m \times m}$ is positive semi-definite, then for any matrix $A \in \mathbb{R}^{n \times m}$, $A B A^\top$ is also positive semi-definite.
\end{lemma}

\begin{lemma}
    \label{op_norm_diff_inequality}
    If matrices $A, B$, and $A - B$ are positive semi-definite, then $\|A - B\|_{\text{op}} \leq \|A\|_{\text{op}}$.
\end{lemma}
% \begin{proof}
%     By Lemma~\ref{op_is_max_lambda_for_psd}, $\|A - B\|_{\text{op}} = \lambda_{\max}(A - B)$ and $\|A\|_{\text{op}} = \lambda_{\max}(A)$. Then by Lemma~\ref{max_eigenvalue_of_psd_diff}, 
%     \begin{align*}
%         \|A - B\|_{\text{op}} = \lambda_{\max}(A - B) \geq \lambda_{\max}(A) = \|A\|_{\text{op}}
%     \end{align*}
% \end{proof}
\section{A Review of Linear Dynamical Systems with Gaussian Noise}
\label{sec_lds_review}
We provide a brief review of discrete-time linear dynamical systems (LDS) with Gaussian noise \citep{roweis1999unifying}.
\subsection{Setting}
A discrete-time, autonomous, LDS with Gaussian noise can be described with the following two equations:

\begin{align}
    \label{standard_lds_state}
    \text{(State Evolution)} \;\;\;
    \vec{z}_t = \Gamma \vec{z}_{t - 1} + w_t, \;\;\; 
\end{align}
\begin{align}
    \label{standard_lds_measurement}
    \text{(Measurement Model)} \;\;\;
    y_t = C \vec{z}_t + v_t, \;\;\;
\end{align}
where $\vec{z}_t \in \mathbb{R}^k$ is the (latent) state of the system with noise process $w_t \overset{\text{i.i.d.}}{\sim} \mathcal{N}(\mu_{z}, W)$, $y_t \in \mathbb{R}$ is some measurement that is observable with noise process $v_t \overset{\text{i.i.d.}}{\sim} \mathcal{N}(\mu_y, V)$, and $\Gamma, C$ are constant matrices. 
% Unlike many settings which assume the system parameters are known, we assume that $\Gamma, C$ are unknown and need to be learned (system identification).

\subsection{Steady-State Kalman Filter}
\label{steady_state_kalman_filter}
With knowledge of $\Gamma, C, \mu_z, W, \mu_y, V$, Kalman filtering is the standard approach for predicting $\vec{z}_t$ using previous measurements $y_1,...,y_{t - 1}$, even if $\vec{z}_t$ is not observed.
% Even if $\vec{z}_t$ is not observed, knowledge of $\Gamma, C, \mu_z, W, \mu_y, V$ enables the use of Kalman filtering to predict $\vec{z}_t$ using previous measurements $y_1,...,y_{t - 1}$. 
Namely, the optimal prediction (in the least mean square sense) for $\vec{z}_t$ would be $z_{t|t - 1}$, where $z_{t|j} := \mathbb{E}[\vec{z}_t | \mathcal{F}_j]$ and $\mathcal{F}_j$ is the sigma algebra generated by previous measurements $y_1,...,y_j$. 

A standard assumption is that the LDS given by equations (\ref{standard_lds_state}) and (\ref{standard_lds_measurement}) is observable:
\begin{assumption}
\label{system_is_observable}
    The observability matrix, $\mathcal{O} = \begin{bmatrix} C \\ C \Gamma \\ C \Gamma^2 \\ \vdots \\ C \Gamma^{k-1} \end{bmatrix} \in \mathbb{R}^{k \times k}$, is full rank.
\end{assumption}

Assumption~\ref{system_is_observable} leads to a steady-state solution \citep{Ljung1999, gornet2022stochastic} that is often employed in practice for applications involving numerous time steps \citep{Kailath2000}. 
% The steady-state Kalman filter requires Assumption~\ref{system_is_observable} to be met.
For simplicity, let $\Tilde{z}_t := z_{t | t - 1}$. The steady-state Kalman filter is as follows, where the prediction and measurement update steps are combined:

\begin{align}
    \label{eqn_steady_state_kalman_filter}
    \Tilde{z}_t = \Gamma \Tilde{z}_{t - 1} + \mu_z + \Gamma K (y_{t - 1} - C \Tilde{z}_{t - 1} - \mu_y)
\end{align}
\begin{align}
    K = PC^\top(CPC^\top + V)^{-1}
\end{align}
\begin{align}
    P = \Gamma P \Gamma^\top + W - \Gamma P C^\top (CPC^\top + V)^{-1}CP\Gamma^\top
\end{align}

% \ziping{Need to help people read and understand these equations. Otherwise, they will get lost.}
% \asim{Asim, could you please check the following section for clarity?}
The prediction of $\Tilde{z}_t$ is recursively computed given the prediction from the previous time-step $\Tilde{z}_{t - 1}$ and the most recent measurement $y_{t - 1}$.
$K$ is the steady-state Kalman gain matrix which acts as a weighting factor balancing the model's predictions
% via equation \ref{eqn_steady_state_kalman_filter} 
with the discrepancy between the model's most recent prediction and measurement.
The update for $K$ is one that yields the minimum mean-square error estimate in the limit.
$P$ is the steady-state version of $P_t := \text{cov}(\vec{z}_t - z_{t|t})$, the error covariance for $\tilde{z}_t$.
Assumption~\ref{system_is_observable} ensures that in the limit, $P_t$ converges to some $P$, which implies the Kalman gain matrix converges to some $K$ \citep{Ljung1999, gornet2022stochastic}.

\section{Discussion on Regret}
\label{app_regret_disc}
In standard stationary bandit settings, one often proves regret bounds with respect to a ``standard oracle" that knows the true fixed reward means $\mu(a)$ for each action and selects the optimal action $a^* = \arg \max_{a \in \mathcal{A}} \mu(a)$. In non-stationary settings, where the mean rewards change over time, many works compare to an equivalent oracle called the dynamic oracle \citep{besbes2014stochastic}. The dynamic oracle is one that knows the true reward means $\mu_t(a)$ at every time step $t$ for each action and selects optimal action at every time step $t$, $a_t^* = \arg \max_{a \in \mathcal{A}} \mu_t(a)$.

% \paragraph{Dynamic Regret} Dynamic regret \cite{besbes2014stochastic} is defined with respect to a dynamic oracle that follows the optimal dynamic sequence of actions. Namely, the dynamic oracle knows the optimal action at every time step $t$ and therefore achieves the largest expected reward at every time step $t$. This is akin to the ``standard oracle" defined above that observes all information in the environment and then acts optimally with that information. \citet{besbes2014stochastic} was able to achieve sub-linear regret because their setting assumes a finite constant (variation budget) of how much the mean rewards can change over time. Their regret bound is with respect to this constant. In our setting, such an oracle would be too strong as we do not make any assumptions on the total variation of the expected rewards over $T$ time steps.

Equivalently, the dynamic oracle is an oracle that observes all information in the environment and then acts optimally with that information. In the latent AR bandit setting, such an oracle knows the ground-truth parameters $\theta^* = [\gamma_0, \gamma_1,...,\gamma_k, \mu_1, \beta_1, ..., \mu_{|\mathcal{A}|}, \beta_{|\mathcal{A}|}, \sigma_z, \sigma_r]$ and observes the latent process (i.e., the realization of $z_t$). 
With access to the ground-truth parameters and the realization of $z_t$, the oracle therefore knows the true reward means $\mu_t(a)$ at every time step $t$ for each action. The oracle selects the optimal action for every $t$: $a_t^* = \arg \max_{a \in \mathcal{A}} \mu_t(a)$.
For a policy $\pi$, the regret is defined as:
%%% don't know if this is equivalent to dynamic oracle so commenting out for now
% \begin{equation*}
%     \text{Regret}(T) = \sum_{t = 1}^T  \underset{a \in \mathcal{A}}{\max} \; \mathbb{E}[r_t(a) | z_t, \theta^*] - \mathbb{E}[r_t(a_t)]
% \end{equation*}
\begin{align*}
    \label{eqn_regret}
    \text{Regret}(T; \pi) =
    \sum_{t = 1}^T  \mathbb{E}[r_t(a_t^*) - r_t(a_t) | \vec{z}_t]
\end{align*}
where $a_t$ is the action selected by the algorithm at time step $t$ following policy $\pi$. The first term is the mean reward obtained by the oracle (i.e., reward obtained by selecting the most optimal action for that time-step) and the second term denotes the mean reward obtained by the agent (where the agent only has information from the history up to but not including time step $t$). 
With no other assumptions, it is impossible to achieve sub-linear regret with respect to this oracle in the latent AR setting.

\subsection{Sub-linear Dynamic Regret}
\label{sublinear_regret_possible}
Sub-linear regret with respect to the dynamic oracle is only possible in environments with vanishing non-stationarity (i.e., there is a budget for the non-stationarity that is sub-linear in $T$). For example in
\citet{besbes2014stochastic}, the non-stationarity is formulated by arbitrary changes to the mean rewards and they assume a finite variation budget $V_T$ of how much the mean rewards can change over time. 
The regret bound for their method Rexp3 is on the order of $V_T^{1/3} T^{2/3}$.
% where they assume $V_t \leq \frac{T}{|\mathcal{A}|}$. 
If $V_T$ scales linearly with $T$, the regret of their method would be linear in $T$.
Similarly in \citet{garivier2011upper}, they assume a finite number of changes to the mean reward or breakpoints $\Upsilon_T$.
The regret bound for their methods discounted UCB and sliding window UCB is on the order of $\sqrt{T \Upsilon_T} \log T$, where $\Upsilon_T = \mathcal{O}(T^{\beta})$ for some $\beta \in [0, 1)$.
If $\Upsilon_T$ scales linearly with $T$, then the regret for their approaches would also achieve linear regret.
In our setting, $\sigma_z^2$, the noise variance on the latent state process, is the mechanism that controls the non-stationarity. We have shown in our main regret bound result (Theorem~\ref{thm_regret}) that for fixed $T$, in environments where $\sigma_{\max}(\Gamma) \leq 1 - \epsilon$ for $\epsilon > 0$ and $\sigma_z^2 = T^{c - 2}$ for some constant $c < 2$, our algorithm achieves sub-linear regret.

\section{Additional Experiments}
\label{app:additional_exps}
\subsection{Verifying the Bias Variance Trade-off}
\label{bias_var_exp}
\begin{figure*}[ht]
  \centering
  \subfigure[]{
    \includegraphics[width=0.45\textwidth]{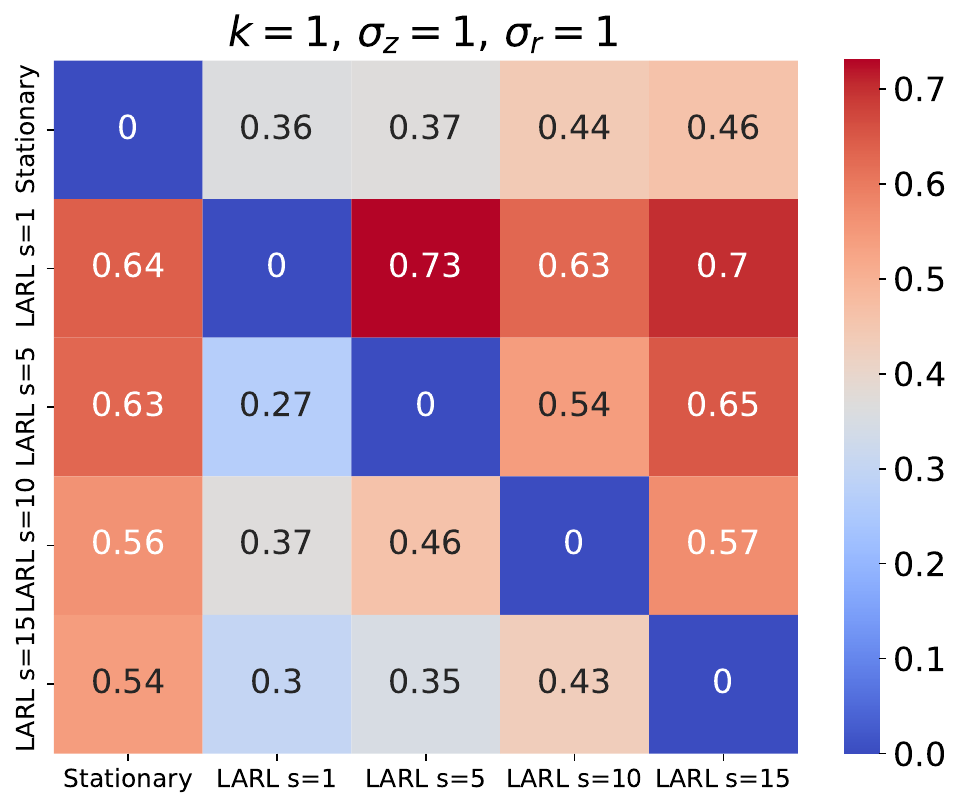}
  }
  \subfigure[]{
    \includegraphics[width=0.45\textwidth]{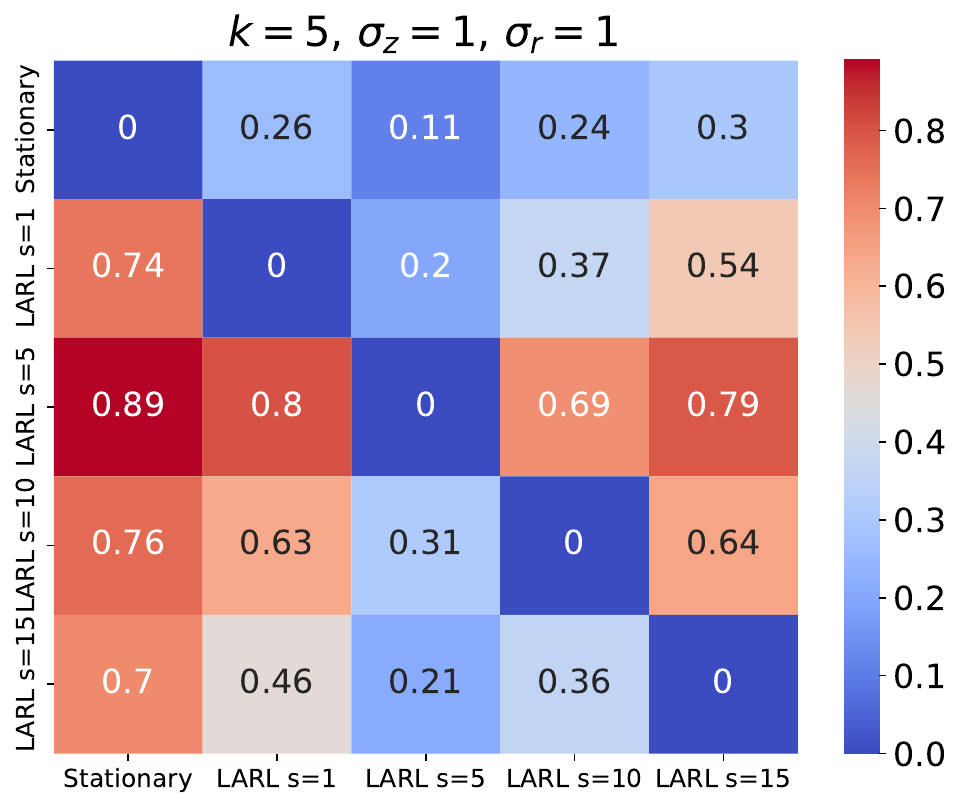}
  }
  \subfigure[]{
    \includegraphics[width=0.45\textwidth]{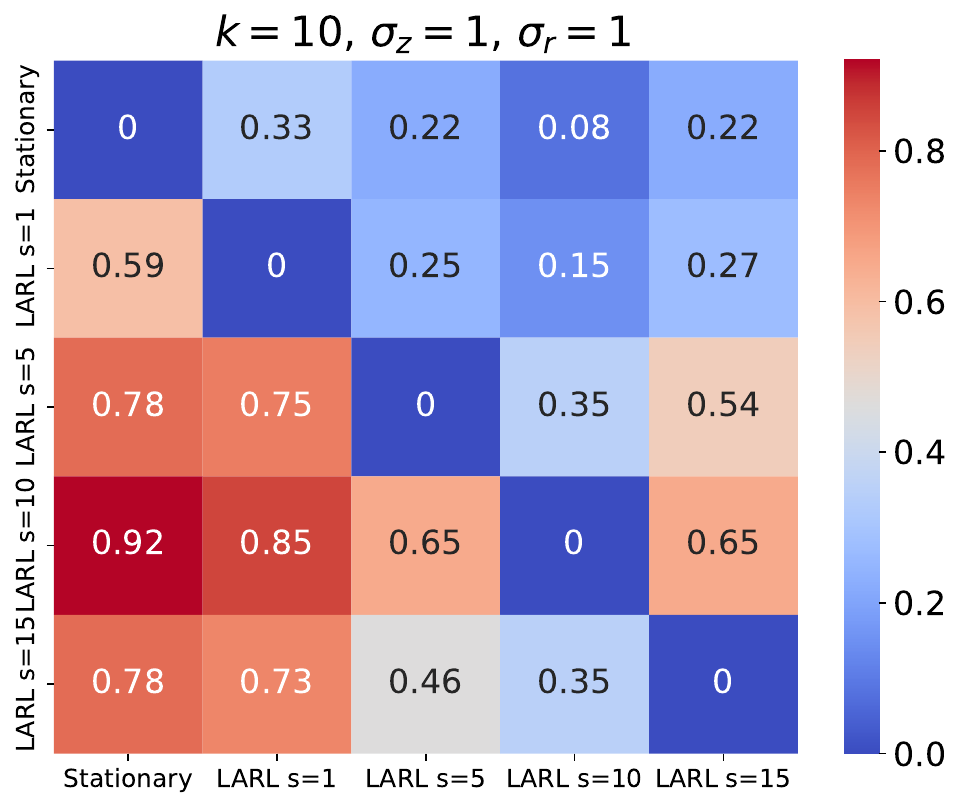}
  }
    \caption{Pairwise comparisons between algorithms in the environment variants where $k = 1, 5, 10$, respectively. Each cell shows the proportion of 100 Monte-Carlo repetitions where the algorithm listed in the row achieved lower cumulative regret than the algorithm listed in the column. 
    % ``Avg.'' shows the average value of each row.
    Even when $s$ is not specifically tuned, our algorithm still outperforms Stationary.
    }
    \label{pairwise_comparisons_1}
\end{figure*}
\begin{figure*}[t]
  \centering
  \subfigure[]{
    \includegraphics[width=0.31\textwidth]{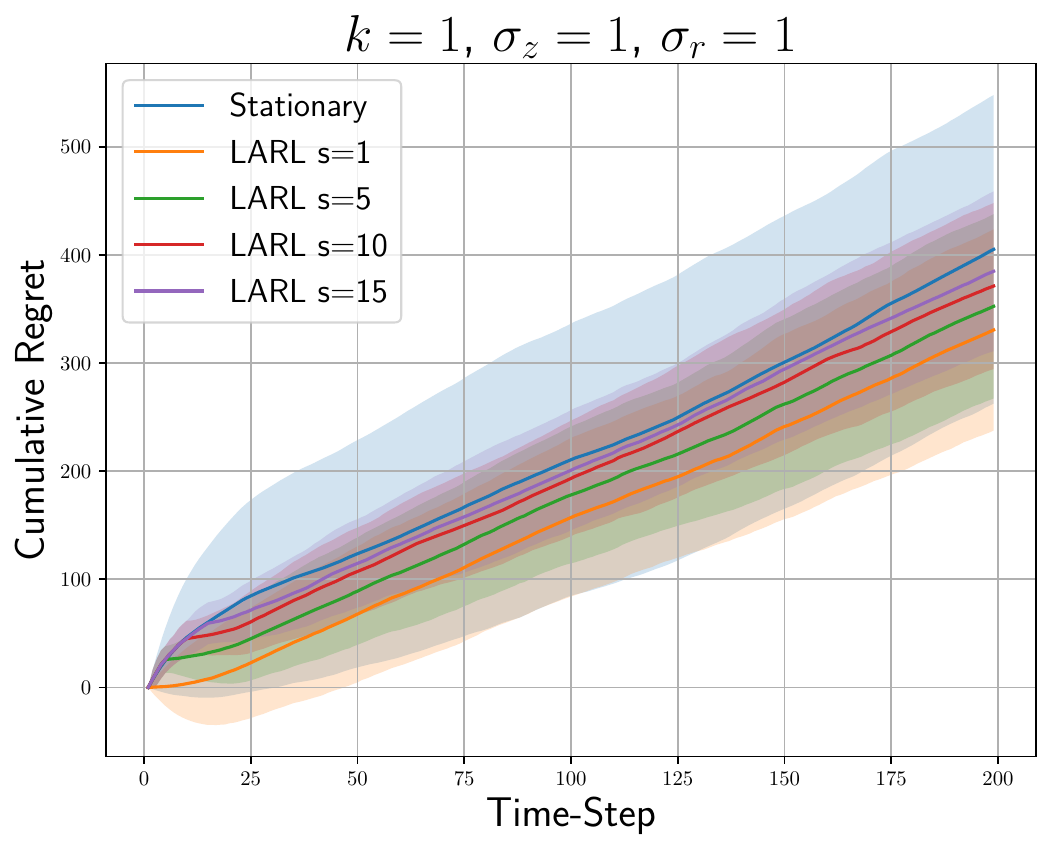}
  }
  \subfigure[]{
    \includegraphics[width=0.31\textwidth]{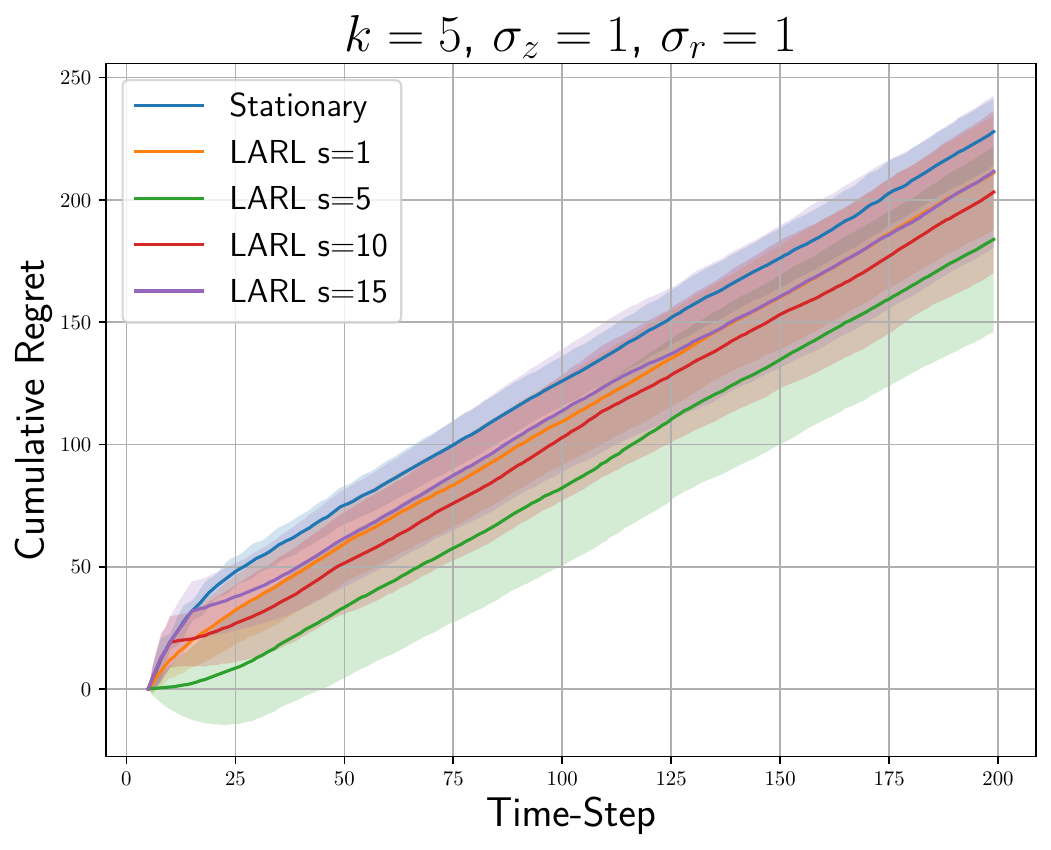}
  }
  \subfigure[]{
    \includegraphics[width=0.31\textwidth]{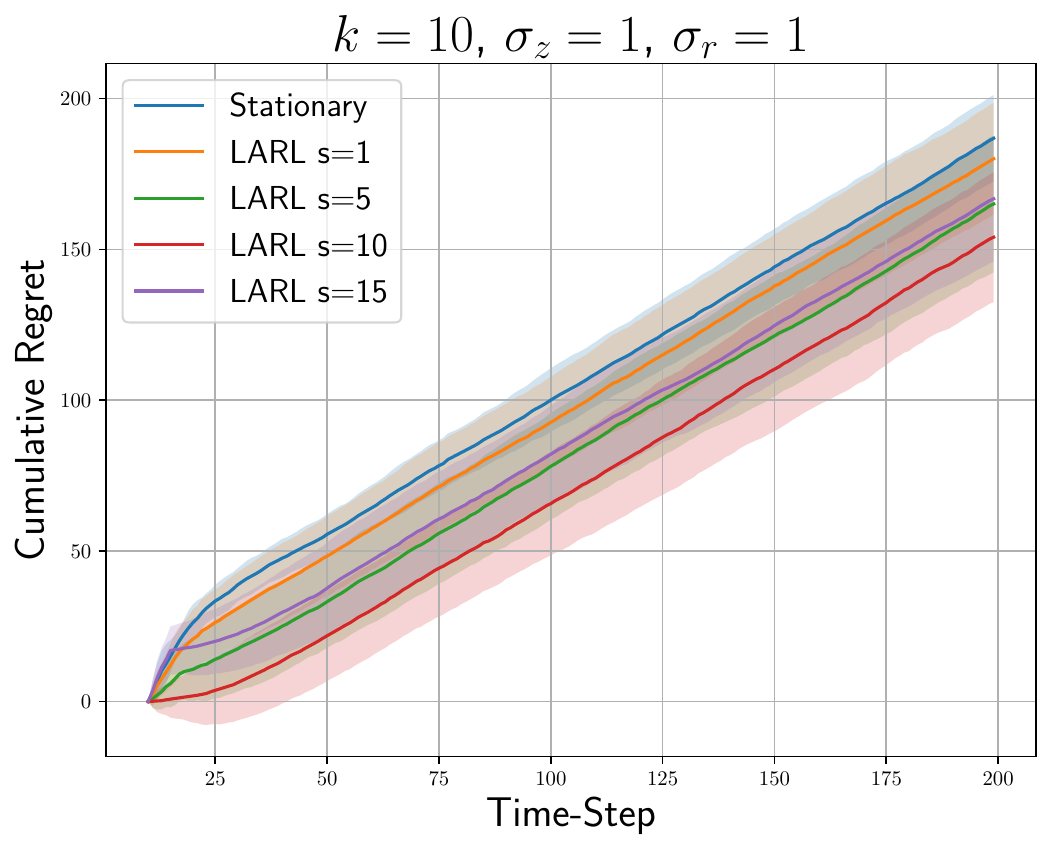}
  }
  \caption{Cumulative regret (Equation~\ref{eqn_regret}) over time with varying choices of $s$ for our algorithm Latent AR LinUCB (Algorithm~\ref{alg_latent_ar_ucb}). For a poor choice of $s$ (either too small or too large compared to $k$) however, our algorithm performs similarly to the stationary. 
  If $s$ is too small, the reward model is under-parameterized.
  % our algorithm would require more data to learn effectively.
  If $s$ is too large, the reward model is over-parameterized.
  % our algorithm's reward model is over-parameterized.
  Line is the average and shaded region is $\pm$ standard deviation across Monte Carlo simulated trials.}
  \label{fig:misspecified_cum_regret}
\end{figure*}

We verify the trade-off with the choice of $s$ as depicted in Theorem~\ref{thm_regret}.
Recall that hyperparameter $s > 0$ dictates the number of recent time steps of history to incorporate into the context. We consider three environment variants where $k = 1, 5, 10$. In each environment variant, we run our algorithm with four different values of $s$, where $s = 1, 5, 10, 15$, compared to ``Stationary'', standard UCB which treats the environment as a stationary multi-armed bandit.

% we perform a sensitivity analysis by varying $s$.
Figure~\ref{pairwise_comparisons_1} shows pairwise comparisons between Stationary and our algorithm LARL with various choices of $s$.
We look at the proportion of times over 100 Monte-Carlo repetitions where the algorithm listed in the row achieved lower total cumulative regret than the algorithm listed in the column. 
% The final column shows the average value of each row (i.e., average performance of each algorithm).
Figure~\ref{fig:misspecified_cum_regret} shows cumulative regret over time.
Even when $s$ is not specifically tuned, our algorithm still outperforms Stationary; however the choice of $s$ does dictate how much our algorithm excels. These simulations verify the bias-variance trade off with the choice of $s$ as shown in the regret bound (Theorem~\ref{thm_regret}). 
% If $s$ is chosen too large or too small, our algorithm is comparable to Stationary. 
%%% ANNA RAMBLINGS %%%
% There is a trade-off. $s$ large means we include more time-steps into the context (recall in comparison, a Kalman filter uses the whole history for the current prediction), but because our algorithm needs to learn its parameters, large $s$ burdens our method with learning more parameters. 
If $s$ is chosen too large, the bias term is small but our algorithm's reward model is burdened with learning many parameters. If $s$ is chosen too small, the bias term is large and our algorithm does not include enough history to inform the current prediction.

\end{document}